\documentclass{article}

\usepackage[margin=1in]{geometry}

\usepackage{microtype}
\usepackage{graphicx}
\usepackage{booktabs} 

\usepackage{hyperref}

\usepackage{amsmath}
\usepackage{amssymb}
\usepackage{bbm}
\usepackage{authblk}

\usepackage{algorithm}
\usepackage{algpseudocode}

\usepackage{natbib} 

\usepackage{xcolor}
\usepackage{subfig}

\newcommand{\orange}[1]{\textcolor{orange}{#1}}

\def\*#1{\boldsymbol{#1}}

\newcommand{\tw}[0]{\textwidth}
\newcommand{\igr}[2]{\includegraphics[clip,width=#1\tw]{#2}}

\newcommand{\pd}[2]{{\frac{\partial #1}{\partial #2}}}
\newcommand{\argmax}{\mathop{\text{argmax}}}

\newcommand{\eq}[1]{(\ref{#1})}

\newcommand{\lw}[1]{\smash{\lower2.ex\hbox{#1}}}

\newcommand{\RR}{\mathbb{R}}

\newcommand{\EE}{\mathbb{E}}

\newcommand{\cN}{{\cal N}}

\newcommand{\cW}{{\cal W}}

\def\UTF#1{--}

\usepackage{mathtools}

\title{Multi-Objective Bayesian Optimization \\with Active Preference Learning}

\author[1]{Ryota Ozaki${}^\dagger$}
\author[1]{Kazuki Ishikawa}
\author[1]{Youhei Kanzaki}
\author[1]{Shinya Suzuki}
\author[3]{Shion Takeno}
\author[2,3]{Ichiro Takeuchi}
\author[1]{Masayuki Karasuyama${}^{\dagger\dagger}$}

\affil[1]{Nagoya Institute of Technology}
\affil[2]{Nagoya University}
\affil[3]{RIKEN AIP}

\affil[ ]{\texttt{${}^\dagger$ ozaki.ryota.mllab.nit@gmail.com, ${}^{\dagger\dagger}$ karasuyama@nitech.ac.jp}}

\date{}

\begin{document}
\maketitle

\begin{abstract}
There are a lot of real-world black-box optimization problems that need to optimize multiple criteria simultaneously. However, in a multi-objective optimization (MOO) problem, identifying the whole Pareto front requires the prohibitive search cost, while in many practical scenarios, the decision maker (DM) only needs a specific solution among the set of the Pareto optimal solutions. We propose a Bayesian optimization (BO) approach to identifying the most preferred solution in the MOO with expensive objective functions, in which a Bayesian preference model of the DM is adaptively estimated by an interactive manner based on the two types of supervisions called the pairwise preference and improvement request. To explore the most preferred solution, we define an acquisition function in which the uncertainty both in the objective functions and the DM preference is incorporated. Further, to minimize the interaction cost with the DM, we also propose an active learning strategy for the preference estimation. We empirically demonstrate the effectiveness of our proposed method through the benchmark function optimization and the hyper-parameter optimization problems for machine learning models.
\end{abstract}

\section{Introduction}
\label{s:introduction}

In many real-world problems, simultaneously optimizing multiple expensive black-box functions $f_1(\*x),\ldots,f_L(\*x)$ are often required.
For example, considering an experimental design for developing novel drugs, there can be several criteria to evaluate drug performance such as the efficacy of the drug and the strength of side effects.
If we pursue high efficacy, the strength of side effects usually tends to deteriorate.
Another example is in the hyper-parameter optimization of a machine learning model, in which multiple different criteria such as accuracy on different classes, computational efficiency, and social fairness should be simultaneously optimized.
%

In general, multi-objective optimization (MOO) problems have multiple optimal solutions (\figurename~\ref{fig:CSF}~(a)), called Pareto optimal solutions, and MOO solvers typically aim to find all of the Pareto optimal solutions.
%
However, the search cost of this approach often becomes prohibitive because the number of Pareto optimal solutions can be large even with small $L$. 
%
On the other hand, in many practical scenarios, a decision maker (DM) only needs one of the optimal solutions that matches their demands (e.g., a drug developer may choose just one drug design considering the best balance between the efficacy and side effects).
Therefore, if we can incorporate the DM's preference at the MOO stage, preferred solutions for the DM can be efficiently identified without enumerating all Pareto solutions.
However, for the DM, directly defining their preference as specific values 
can be difficult.

We propose a Bayesian optimization (BO) method for the preference-based MOO, optimizing $\*x$ through an interactive (human-in-the-loop based) estimation of the preference based on weak-supervisions provided by the DM.
Let $U: \RR^L \rightarrow \RR$ be a utility function that quantifies the DM preference.
We assume that when the DM prefers $\*f \in \RR^L$ to $\*f^\prime \in \RR^L$, the utility function values satisfy 
$U(\*f) > U(\*f^\prime)$.
By using $U$, the optimization problem can be formulated as 
$\max_{\*x} U(\*f(\*x))$, 
where 
$\*f(\*x) = (f_1(\*x), \ldots, f_L(\*x))^\top$.  
%
Our proposed method is based on a Bayesian modeling of $\*f(\*x)$ and $U$, by which uncertainty of both the objective functions and the DM preference can be incorporated.
%
For $\*f(\*x)$, we use the Gaussian process (GP) regression, following the standard convention of BO.
For $U$, we employ a Chebyshev scalarization function (CSF) based parametrized utility function because of its simplicity and capability of identifying any Pareto optimal points depending on a setting of the preference parameter $\*w \in \RR^L$ (\figurename~\ref{fig:CSF}~(b)). 
%

%
%
%
%

%
%

To estimate utility function $U$ (i.e., to estimate the parameter $\*w$), we consider two types of weak supervisions provided by the interaction with the DM.
%
First, we use the pairwise comparison (PC) over two vectors $\*f, \*f^\prime \in \RR^L$.
In this supervision, the DM answers whether $U(\*f) > U(\*f^\prime)$ holds based on their preference. 
%
%
In the context of preference learning \citep[e.g.,][]{pregp}, this first type of supervision is widely known that the relative comparison is often much easier to be provided by the DM than the exact value of $U(\*f)$.
%
As the second preference information, we propose to use improvement request (IR) for a given $\*f \in \RR^L$.
The DM provides the index $\ell$ for which the DM hopes that the $\ell$-th dimension of $\*f$, i.e., $f_{\ell}$, is required to improve most among all the $L$ dimensions.
To our knowledge, this way of supervision has never been studied in preference learning nevertheless it is obviously easy to provide and important information for the DM. 
We show that IR can be formulated as a weak supervision of the gradient of the utility function.

%
We show that the well-known expected improvement (EI) acquisition function of BO can be defined for the optimization problem
$\max_{\*x} U(\*f(\*x))$.
%
Here, the expectation is taken over the both of $U$ and $\*f(\*x)$, by which optimization is performed based on the current uncertainty for both of them.
%
Further, to reduce querying cost to the DM, we also propose active learning that selects effective queries (PC or IR) to estimate the preference parameter $\*w$.
Our contributions can be summarized as follows:
\begin{itemize}
 \item 
       We propose a preference-based MOO algorithm by combining BO and preference learning in which both the multi-objective functions $\*f(\*x)$ and the DM preference (utility function) are modeled by a Bayesian manner.
 \item Bayesian preference learning for the utility function is proposed based on two types of weak supervisions, i.e., PC and IR.
       In particular, IR is a novel paradigm of preference learning.
       %
       %
 \item Active learning for PC and IR is also proposed.
       Our approach is based on BALD (Bayesian Active Learning by Disagreement) \citep{houlsby2011bayesian}, which uses mutual information to measure the benefit of querying.
 \item As an extension, we also introduce a GP based utility function for cases that higher flexibility is needed to capture the preference of the DM.
       Since the utility function for MOO should be monotonically non-decreasing, we combine a GP with monotonicity information \citep{monotonicGPR} with preferential learning \citep{pregp}.
       %
 \item Numerical experiments on several benchmark functions and hyperparameter optimization of machine learning models show superior performance of our framework compared with baselines such as an MOO extension of BO without preference information.
\end{itemize}


\section{Problem Setting}
\label{s:problem-setting}

\begin{figure}
 \begin{center}
  \igr{.55}{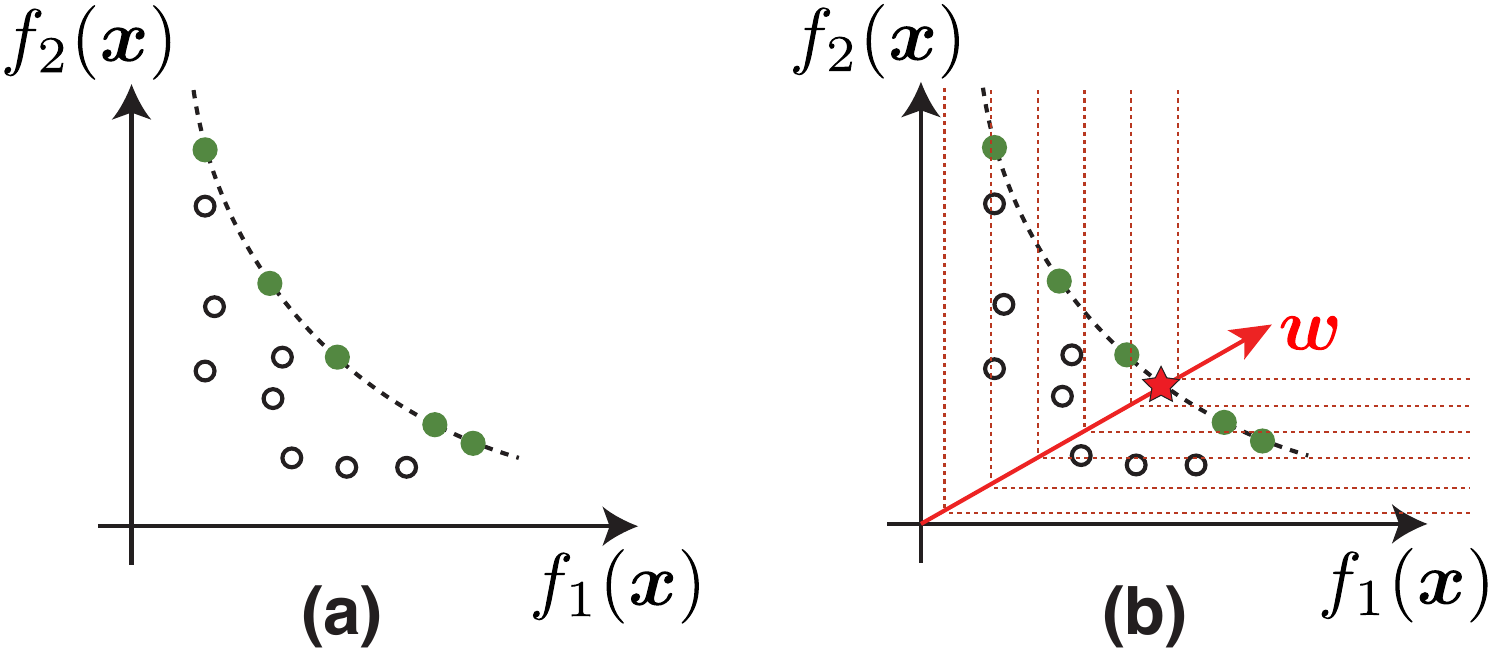}
 \end{center}
 \caption{
 (a) Illustration of Pareto optimal points. 
 The green points are Pareto optimal because, from each one of the green points, no points can improve both objectives simultaneously. 
 The dashed line is an underlying set of all the Pareto optimal points, called the Pareto front.
 (b) Illustration of CSF (see \S~\ref{ss:preference-model} for detail of CSF).
 The direction of the red arrow corresponds to $\*w$.
 The red dashed lines are the contour lines of CSF, by which the red star becomes the optimal point under the DM's preference.
 }
 \label{fig:CSF}
\end{figure}

We consider a multi-objective optimization (MOO) problem that maximizes $L$ objective functions.
Let 
$f_\ell(\*x)$ for $\ell \in [L]$
be a set of objective functions
and 
$\*f(\*x) = (f_1(\*x), \ldots, f_L(\*x))^\top$ 
be their vector representation, where $\*x \in \RR^d$ is an input vector.
%
%
%
In general, an MOO problem can have multiple optimal solutions, called the Pareto optimal points. 
For example, in \figurename~\ref{fig:CSF}~(a), all the green points are optimal points.
See MOO literature \citep[e.g.,][]{chebyshev} for the detailed definition of the Pareto optimality.

Since the number of the Pareto optimal points can be large even with small $L$ \cite[e.g.,][]{ishibuchi2008evolutionary}, identifying all the Pareto optimal points can be computationally prohibitive. 
Instead, we consider optimizing a {\it utility function}
$U: \RR^L \rightarrow \RR$ that satisfies
$U(\*f) > U(\*f^\prime)$
when the DM prefers $\*f$ to $\*f^\prime$, where $\*f, \*f^\prime \in \RR^L$.
The optimization problem that seeks the solution preferred by the DM is represented as
\begin{align}
 \max_{\*x} \ U(\*{f}(\*x)).
 \label{eq:U_w}
\end{align}
In our problem setting, both the functions $\*f(\*x)$ and $U(\*f)$ are assumed to be unknown.

About the querying to the objective function $\*f(\*x)$, we follow the standard setting of BO.
Let 
$\*y_i = \*f(\*x_i) + \*\varepsilon$ 
be a noisy observation of
$\*f(\*x_i)$
where
$\*\varepsilon \sim \cN(0, \sigma^2 I)$
is an independent noise term with the variance $\sigma^2$ and the identity matrix $I$. 
Since observing $\*y_i$ requires high observation cost, a sample efficient optimization strategy is required. 

On the other hand, directly observing the value of the utility function 
$U(\*{f}(\*x))$
is usually difficult because of difficulty in defining a numerical score of the DM preference.
Instead, we assume that we can query to the DM about the following two types of weak supervisions:
\begin{description}
 \item[Pairwise Comparison (PC):] 
	    PC indicates the relative preference over given two $\*f$ and $\*f^{\prime}$, i.e., the DM provides whether $\*f$ is preferred than $\*f^{\prime}$ or not.
	    %
 \item[Improvement Request (IR):] 
	    IR indicates the dimension $\ell$ in a given $\*f$ that the DM considers improvement is required most among $\ell \in [L]$.
\end{description}
For the DM, these two types of information are much easier to provide than the exact value of $U(\*f)$ itself.
PC is a well-known format of a supervision in the context of preference learning \citep{pregp} and dueling bandit \citep{sui2018advancements}.
On the other hand, it should be noted that IR has not been studied in these contexts to our knowledge, though it considers a practically quite common scenario. 
%
%


\section{Proposed Method}
\label{s:proposed-method}


In this section, we first describe the modeling of the MOO objective functions $\*f(\*x)$ in \S~\ref{ss:GP-for-objective}. 
Next, in \S~\ref{ss:preference-model}, the modeling of the utility functions $U$ that represents the DM preference is described. 
In \S~\ref{ss:acqs}, we show the acquisition functions for BO of $U(\*f(\*x))$ and active learning of $U$.
Then, the entire algorithm of the proposed method is shown in \S~\ref{ss:algorithm}.
Finally, we discuss a variation of the utility function in \S~\ref{s:discussion-utility}.

\subsection{Gaussian Process for Objective Functions}
\label{ss:GP-for-objective}

As a surrogate model of $\*f(\*x)$, we employ the Gaussian process (GP) regression.
For simplicity, the $L$ dimensional-output of $\*f(\*x)$ is modeled by the $L$ independent GPs, each one of which has 
$k(\*x,\*x^\prime)$ 
as a kernel function.
%
%
When we observe a set of $t$ observations
$D_{\rm GP} = \{(\*x_i, \*y_i)\}_{i=1}^t$, 
the predictive distribution can be obtained as the posterior 
$p(\*f(\*x) \mid D_{\rm GP})$,
for which the well-known analytical calculation is available \citep[see e.g.,][]{gp}.  
Although we here use the independent setting for the $L$ outputs, incorporating correlation among them is also possible by using the multi-output GP \citep{kernel}. 
%

\subsection{Bayesian Modeling for Utility Function}
\label{ss:preference-model}

\begin{figure}
 \begin{center}
  \igr{.55}{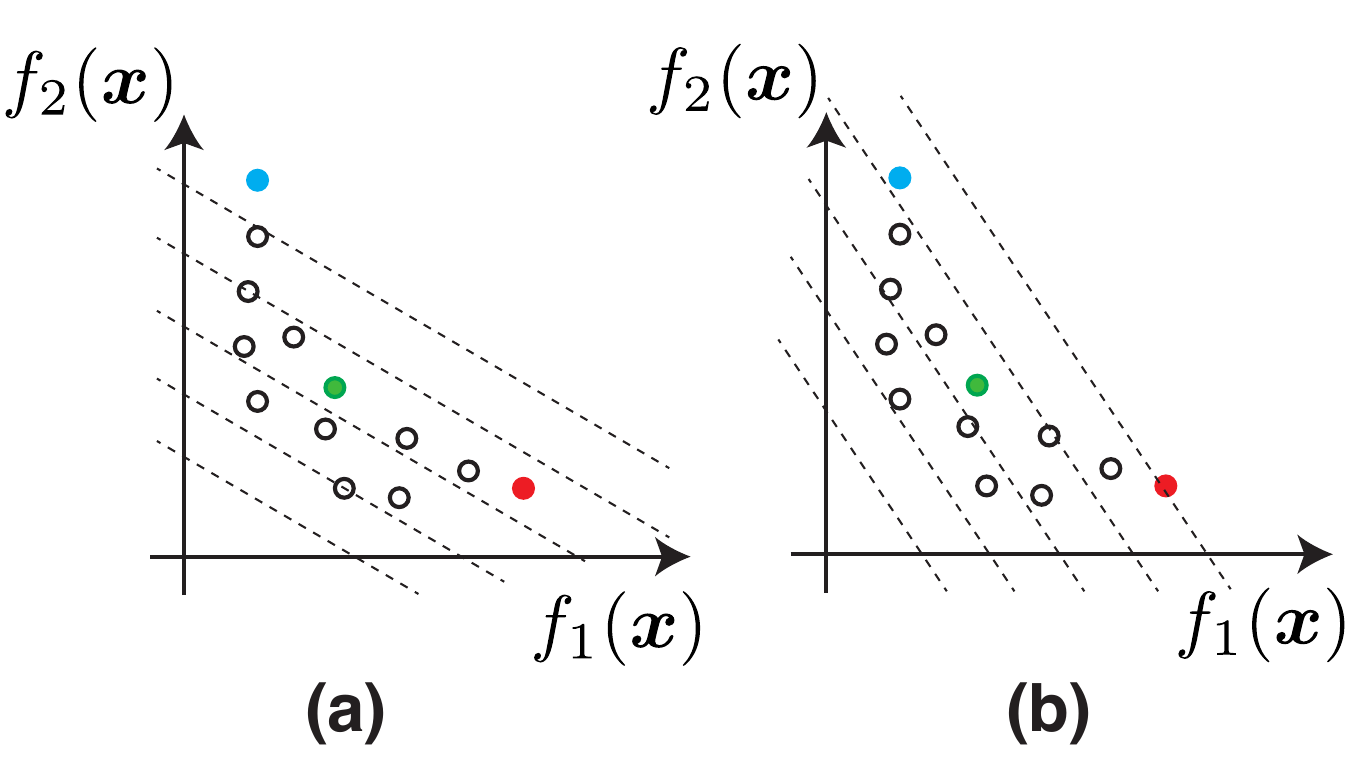}
 \end{center}
 \caption{
 Illustrative examples of contour plots the linear utility function
 $U(\*f) = \sum_{\ell=1}^L w_\ell f_\ell$, where $w_\ell \geq 0$.
 The plots (a) and (b) have different coefficients.
 In (a), the blue point is the optimal in a sense of $U$, while the red point is the optimal in (b).
 On the other hand, the green point can never be the optimal point by any selection of the coefficients.
 }
 \label{fig:linear-utility}
\end{figure}

%
The simplest choice of the utility function is a linear function 
$U(\*f) = \sum_{\ell=1}^L w_\ell f_\ell$ 
with a weight parameter $w_\ell \geq 0$.
However, in a linear utility function, depending of the shape of the Pareto front, it is possible that: 1) there exists a Pareto optimal solution that cannot be identified by any weight parameters, and 2) only one objective function with the highest weight is exclusively optimized (\figurename~\ref{fig:linear-utility}).
To incorporate the DM preference, variety of scalarization functions have been studied \citep{bechikh2015preference}. 
We here mainly consider the following Chebyshev scalarization function (CSF) \citep{chebyshev}: 
\begin{align}
 U(\*{f}(\*x))=\min\left(\frac{f_1(\*x)}{w_1},\ldots,\frac{f_L(\*x)}{w_L}\right), 
 \label{eq:Cheby}
\end{align}
where 
$\*w = (w_1, \ldots, w_L)^\top$
is a preference vector that satisfies 
$\sum_{\ell \in [L]} w_\ell = 1$ and $w_\ell > 0$.
\figurename~\ref{fig:CSF}~(b) is an illustration of CSF.
In the MOO literature \citep{chebyshev}, it is known that for any Pareto optimal solution $\*f^\star$, there exist a weighting vector $\*w$ under which the maximizer $\*x$ of \eqref{eq:Cheby} derives $\*f^\star = \*f(\*x)$.
For further discussion on the choice of the utility function, see \S~\ref{s:discussion-utility}.


\subsubsection{Likelihood for Pairwise Comparison}
\label{sss:likelihood-outcome}

We define the likelihood of PC, inspired by the existing work on preference learning of a GP \citep{pregp}.
%
Suppose that the DM prefers $\*f^{(i)}$ to $\*f^{(i^\prime)}$, for which we write  
$\*f^{(i)} \succ \*f^{(i^\prime)}$.
We assume that the preference observation is generated from the underlying true utility $U$ contaminated with the Gaussian noise as follows:
\begin{align*}
 \*f^{(i)} \succ \*f^{(i^\prime)}
 \Leftrightarrow 
 U(\*f^{(i)}) + \epsilon_i > U(\*f^{(i^\prime)}) + \epsilon_i^{\prime},
\end{align*}
where 
$\epsilon_i, \epsilon_i^{\prime} \sim N(0, \sigma_{\rm PC}^2)$
is the Gaussian noise having the variance $\sigma_{\rm PC}^2$.
For a given set of $n$ preference observations 
$\{ \*f_i \succ \*f^{\prime}_i \}_{i=1}^{n}$, 
the likelihood is written as 
\begin{align} 
 p(\{ \*f^{(i)} \succ \*f^{(i^\prime)} \}_{i=1}^{n} | \*{w})
 \! = \!
 \prod_{i=1}^{n} \Phi\left(\frac{U(\*f^{(i)}) - U(\*f^{(i^\prime)})}{\sqrt{ 2 } \sigma_{\rm PC} } \right),
 \label{eq:likelihood-PC}
\end{align}
where 
$\Phi$
is the cumulative distribution function of the standard Gaussian distribution.

\subsubsection{Likelihood for Improvement Request}
\label{sss:likelihood-improve}

For a given 
$\*f^{(i)} = (f^{(i)}_{1}, \ldots, f^{(i)}_{L})^\top$, 
if the DM considers 
$f^{(i)}_{\ell_i}$ 
has a higher priority to be improved more than other
$f^{(i)}_{\ell^\prime_i}$, 
we write
$\ell_i \succ \ell_i^\prime$.
In IR, an observation is a dimension 
$\ell_i$ 
for which the DM requires improvement most strongly among $L$ dimensions.
This can be considered that we observe $L - 1$ relations 
$\ell_i \succ \ell_i^\prime$
for $\ell_i^\prime \in [L] \setminus \ell_i$.

\begin{figure}[t]
 \begin{center}
  \igr{.24}{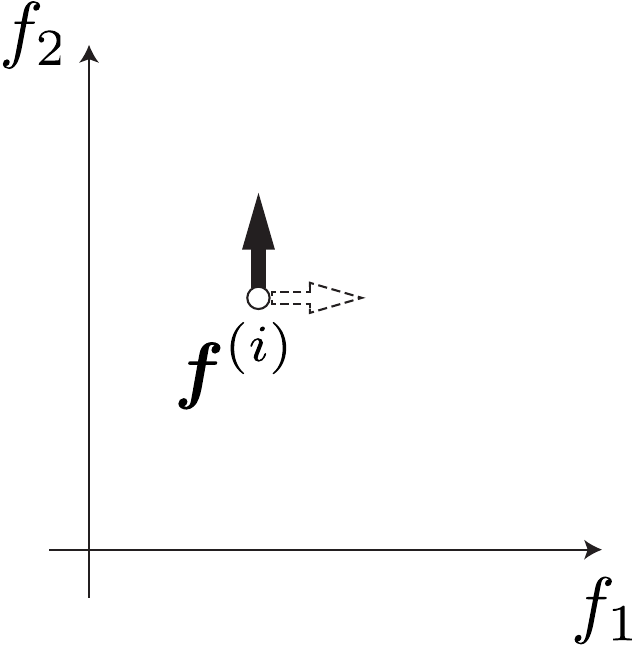}
  \caption{
  Assume that the DM requests that the direction of $f_2$ should be improved than $f_1$ at $\*f^{(i)}$. 
  This indicates that the utility function should have larger increase when $\*f^{(i)}$ increases along with the axis of $f_2$ (the direction of the black arrow) compared with the axis of $f_1$ (the white arrow).
  }
  \label{fig:IR}
 \end{center}
\end{figure}

The observation 
$\ell_i \succ \ell_i^\prime$ for $\*f^{(i)}$
can be interpreted that the gradient of $U(\*f^{(i)})$ with respect to 
$f^{(i)}_{\ell_i}$ 
is larger than those of
$f^{(i)}_{\ell^\prime_i}$.
In other words, the direction of 
$f^{(i)}_{\ell_i}$ 
should improve the preference
$U(\*f^{(i)})$ 
more rapidly than the direction of 
$f^{(i)}_{\ell^\prime_i}$ (\figurename~\ref{fig:IR}).
Let
$g_{\ell}(\*f) = \pd{U(\*f)}{f_{\ell}}$
be the $\ell$-th dimension of the gradient of $U(\*f)$.
Then, the event
$\ell_i \succ \ell_i^{\prime}$
is characterized through the underlying gradient 
$g_{\ell}(\*f)$ 
as follows:
\begin{align*}
 %
 \ell_i \succ \ell_i^{\prime}
 \Leftrightarrow 
 g_{\ell_i}(\*f^{(i)}) - g_{\ell_i^\prime}(\*f^{(i)}) > e_i
\end{align*}
where 
$e_i \sim N(0, \sigma_{\rm IR}^2)$ 
is the observation noise with the variance $\sigma_{\rm IR}^2$.
Suppose that we have $m$ observations 
$\{\ell_{i} \succ \ell^{\prime}_{i}\}_{i=1}^{m}$ 
in total.
The likelihood is 
\begin{align}
 p(\{\ell_{i} \succ \ell^{\prime}_{i}\}_{i=1}^{m} \mid \*w)
 = \prod_{i=1}^{m} \Phi\left(\frac{ g_{\ell_i}(\*f^{(i)})) - g_{\ell_i^\prime}(\*f^{(i)}) }{ \sigma_{\rm IR} } \right).
 \label{eq:likelihood-IR}
\end{align}

\subsubsection{Prior and Posterior}
\label{sss:prior-posterior}

We employ the Dirichlet distribution 
$p(\*w)=\frac{1}{B(\*\alpha)} \prod_{i=1}^{L} w_{i}^{\alpha_{i}-1}$
as a prior distribution of $\*w$ because it has the constraint $\|\*w\|_1=1$, where $B$ is the beta function and 
$\*\alpha = [\alpha_1,\ldots,\alpha_L]^\top$
is a parameter.
Let $D_{\mathrm{pre}}$ be a set of PC and IR observations. 
The posterior distribution of $\*w$ is written as
\begin{align}
 \begin{split}  
 & p(\*w \mid D_{\mathrm{pre}}) \propto p(\*w) 
 \\
 & \quad \times 
  p(\{ \*f^{(i)} \succ \*f^{(i^\prime)}\}_{i=1}^{n} \mid \*w) 
 \
 p(\{\ell_{j} \succ \ell^{\prime}_{j}\}_{j=1}^{m} \mid \*w). 
 \end{split}
 \label{eq:w-posterior}
\end{align}
%



\subsection{Acquisition Functions}
\label{ss:acqs}


\subsubsection{Bayesian Optimization of Utility Function}
\label{sss:acqs}

Our purpose is to identify $\*x$ that maximizes $U(\*f(\*x))$, for which we employ the well-known expected improvement (EI) criterion.
%
Let
$U_{\rm best} = \max_{i \in [t]} U(\*y_i)$
be the best utility function value among already observed 
$\*y_i$.
%
The acquisition function that selects the next $\*x$ is defined by
\begin{align}
 \alpha_{\rm EI}(\*x) = 
 \EE_{\*f(\*x), \*w}\left[ \max \left\{
 U \left( \*f(\*x) \right) 
 - 
U_{\rm best}, 
 0
 \right\} \right].
 \label{eq:EI}
\end{align}
Unlike the standard EI in BO, the expectation is jointly taken over $\*f(\*x)$ and $\*w$, and the current best term 
$U_{\rm best}$
is also random variable (depending on $\*w$).
Therefore, the analytical calculation of \eq{eq:EI} is difficult and we employ the Monte Carlo (MC) method to evaluate \eq{eq:EI}.
The objective function $\*f(\*x)$ can be easily sampled because it is represented by the GPs (Note that another possible approach for the expectation over $\*f(\*x)$ is to transform it into the expectation over one dimensional space of $U(\*f(\*x))$ on which numerical integration can be more accurate. See Appendix~\ref{app:ei-calculation} for detail).
The parameter of utility function $\*w$ is sampled from the posterior \eq{eq:w-posterior}, for which we use the standard Markov chain MC (MCMC) sampling.
Since both of $\*f(\*x)$ and $U$ are represented as Bayesian models, uncertainty with respect to both the objective functions and the user preference are incorporated in our acquisition function.
%
%

\begin{figure}
 \begin{center}
  \igr{.7}{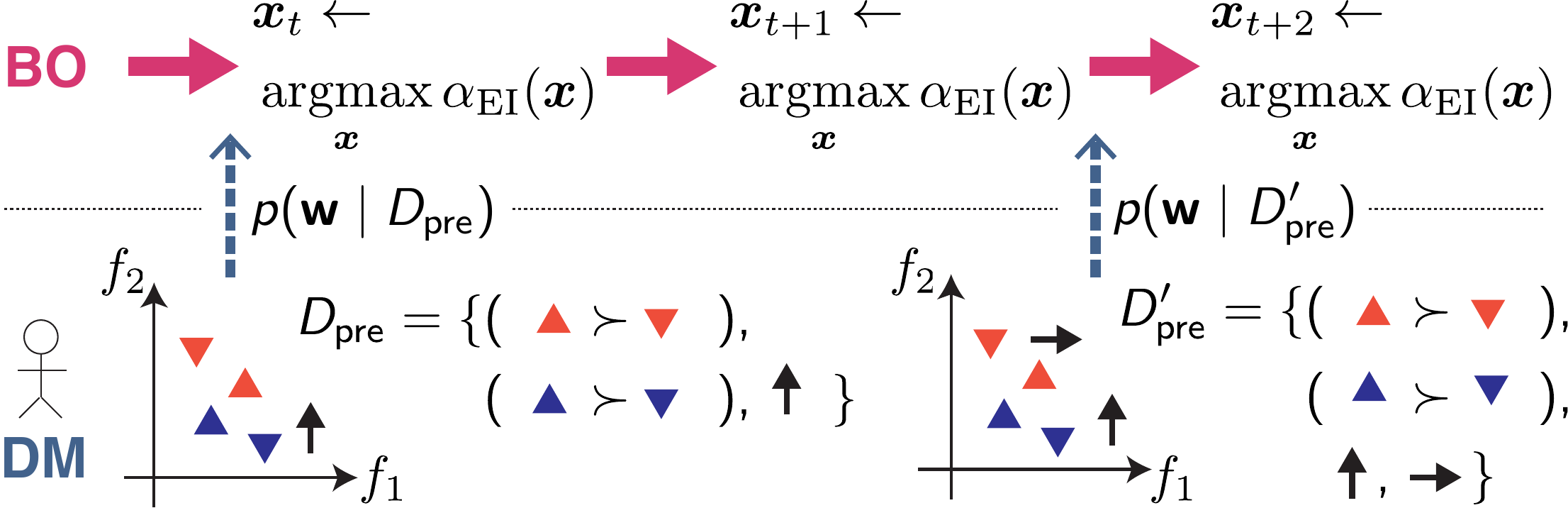}
 \end{center}
 \caption{
 A schematic illustration of the procedure of the proposed framework.
 Let $\*x^{(t)}$ be the selected $\*x$ at the $t$-th iteration of BO.
 The DM can add the preference information at any time point.
 A pair in a PC observation is represented by the two triangles with the same color (e.g., $\orange{\blacktriangle} \succ \orange{\blacktriangledown}$). 
 Each black arrow represents the direction that the DM requests to improve. 
 %
 In $D_{\rm pre}^\prime$, the DM newly adds the request for the improvement on the horizontal direction.
 }
 \label{fig:overview}
\end{figure}

\subsubsection{Active Learning for Utility Function Estimation}
\label{sss:acqs}

We also propose an active learning (AL) acquisition function for efficiently estimating the utility function $U$.
%
%
Since PC- and IR- observations require interactions with the DM, accurate preference estimation with the minimum observations is desired.
Our AL acquisition function is based on the Bayesian active learning framework called BALD (Bayesian Active Learning by Disagreement) \citep{houlsby2011bayesian}.
BALD is an information theoretic approach in which the next query is selected by maximizing mutual information (MI) between an observation and a model parameter. 

We here describe the case of PC only because for IR, almost the same procedure is derived (See Appendix~\ref{app:acq-ir} for detail).
We need to select a pair
$\*f$ 
and 
$\*f^{\prime}$ 
that efficiently reduces the uncertainty of $\*w$.
Let 
\begin{align*}
 z_{\rm PC} = 
 \begin{cases}
  1 & \text{ if } \*f \succ \*f^{\prime}, 
  \\
  0 & \text{ if } \*f^{\prime} \succ \*f
 \end{cases}
\end{align*}
be the indicator of the preference given by the DM.
Our AL acquisition function is defined by the MI between $z_{\rm PC}$ and $\*w$:
\begin{align}
 \mathrm{MI}(z_{\rm PC} ; \*w) = H[ z_{\rm PC} ] - \EE_{\*w}[H[ z_{\rm PC} \mid \*w]],
 \label{eq:MI-PC}
\end{align}
where $H$ represents the entropy.
\citet{houlsby2011bayesian} clarify that this difference of the entropy representation results in a simpler computation than other equivalent representations of MI.

%
For the first term of \eq{eq:MI-PC}, since $z_{\rm PC}$ follows the Bernoulli distribution, we have
\begin{align*}
 H[z_{\rm PC}] 
 = \sum_{z_{\rm PC} \in \{ 1, 0 \}} p(z_{\rm PC}) \log p(z_{\rm PC}).
\end{align*}
Unfortunately, 
$p(z_{\rm PC})$, 
in which $\*w$ is marginalized, is difficult to evaluate analytically.
On the other hand, the conditional distribution
$p(z_{\rm PC} \mid \*w)$
is easy to evaluate as shown in \eq{eq:likelihood-PC}.  
Therefore, we employ a sampling based approximation
\begin{align*}
 p(z_{\rm PC}) \approx \sum_{\*w \in \cW} p(z_{\rm PC} \mid \*w) / |\cW|,
\end{align*}
where $\cW$ is a set of $\*w$ generated from the posterior \eq{eq:w-posterior} (e.g., by using the MCMC sampling).
For the second term of \eq{eq:MI-PC}, the same sample set $\cW$ can be used as
\begin{align*}
 & \EE_{\*w}[H[ z_{\rm PC} \mid \*w]] \approx 
 \sum_{\*w \in \cW} \sum_{z_{\rm PC} \in \{ 1, 0 \}} \frac{p(z_{\rm PC} \mid \*w) \log p(z_{\rm PC} \mid \*w)}{ |\cW| }.
\end{align*}
%


\subsection{Algorithm}
\label{ss:algorithm}

The procedure of the proposed framework is shown in \figurename~\ref{fig:overview} and Algorithm~\ref{alg:proposed-method} (MBO-APL: Multi-objective BO with Active Preference Learning).
Note that although Algorithm~\ref{alg:proposed-method} also only considers the case of PC, the procedure for IR is almost the same (change MI for IR, shown in Appendix~\ref{app:acq-ir}).
BO and the preference learning can be in parallel because the training of the GPs and $\*w$ are independent.
When the DM adds the preference data (PC and/or IR) into $D_{\rm pre}$, the posterior of $\*w$ is updated.
The updated posterior can be immediately used in the next acquisition function calculation of BO (in \figurename~\ref{fig:overview}, for example, $p(\*w \mid D_{\rm pre}^\prime)$ can be used to determine $\*x^{(t+2)}$).

\begin{algorithm}[!t]
    \caption{Proposed Method}\label{alg:proposed-method}
    \begin{algorithmic}[1]
     \Procedure{MBO-APL}{}
     \State Run \textsc{Active-Pref-Learning} in background
     \For{$t = 1, \dots$}
            \State Fit GPs to $D_{\rm GP} = \{(\*x_i, \*y_i)\}_{i=1}^t$
            \State $\*x_{t+1} \gets \argmax_{\*x} \alpha_{\rm EI}(\*x)$ \par
            \hskip\algorithmicindent \qquad using current $p(\*w \mid D_{\rm pre})$
            \State Observe $(\*x_{t+1}, \*y_{t+1})$ \par 
            \hskip\algorithmicindent \qquad and $D_{\rm GP} \gets  D_{\rm GP} \cup (\*x_{t+1}, \*y_{t+1})$
     \EndFor 
     \EndProcedure
     \Procedure{Active-Pref-Learning}{}
        \For{$t = 1, \dots$}
            \State Update $p(\*w \mid D_{\rm pre})$ with the current $D_{\rm pre}$
            \State $\*f, \*f^\prime \gets \argmax_{\*f, \*f^\prime} {\rm MI}(z_{\rm PC}; \*w)$
            \State Query $z_{\rm PC}$ to the DM and add the result to $D_{\rm pre}$
        \EndFor
     \EndProcedure
    \end{algorithmic}
\end{algorithm}

\subsection{Selection on Utility Function}
\label{s:discussion-utility}

Although we mainly focus on \eqref{eq:Cheby} as a simple example of the utility function, different utility functions can be used in our framework.

\subsubsection{Augmented Chebyshev Scalarization Function}

For example, in MOO literature, the following augmented CSF is often used \citep{bechikh2015preference,hakanen2017using}:
\begin{align*}
 U(\*f(\*x)) = 
 \min_{\ell \in [L]} \frac{f_\ell(\*x) - f_\ell^{\rm ref}}{w_{\ell}} 
 + \rho \sum_{\ell \in [L]}
 \frac{f_\ell(\*x) - f_\ell^{\rm ref}}{w_{\ell}},
\end{align*}
where 
$\rho > 0$ 
is an augmentation coefficient (usually a small constant) and 
$f_\ell^{\rm ref}$ 
is a reference point.
If the DM can directly provide the reference point, $f_\ell^{\rm ref}$ is a fixed constant, while it is also possible that we estimate $f_\ell^{\rm ref}$ as a random variable from PC (See Appendix~\ref{app:augmented-csf} for detail). 
%
The second term avoids weakly Pareto optimal solutions.  
Using the augmented CSF in our framework is easy because the posterior of $\*w$ (and $f_\ell^{\rm ref}$) can be derived by the same manner described in \S~\ref{ss:preference-model}. 

\subsubsection{Gaussian Process with Monotonicity Constraint}

Instead of parametric models such as CSF, nonparametric approaches are also applicable to defining the utility function.
Since our problem setting is the maximization of $\*f(\*x)$, 
the utility function
$U$
should be monotonically non-decreasing.
%
Therefore, for example, the GP regression with the monotonicity constraint \citep{monotonicGPR} can be used to build a utility function with high flexibility.

We combine the monotonic GP with the preferential GP by which a monotonically non-decreasing GP regression can be approximately constructed from PC- and IR- observations.
We here only consider PC observations 
$\{ (\*f^{(i)}, \*f^{(i^\prime)})\}_{i=1}^n$, but IR observations can be incorporated based on the same manner because the derivative of a GP is also a GP.
Let
$\*U=\left(U(\*f^{(1)}),\ldots, U(\*f^{(N)})\right)^{\top}$, where $N > n$ and 
$\{ \*f^{(n+1)}, \ldots, \*f^{(N)} \} = \{ \*f^{(1^\prime)}, \ldots, \*f^{(n^\prime)} \} \setminus \{ \*f^{(1)}, \ldots, \*f^{(n)} \}
$
.
%
The monotonicity constraints can be seen as an infinite number of the derivative constraints 
$\pd{U(\*f)}{f_\ell} \geq 0$ 
for 
$\forall \*f = (f_1, \ldots, f_L)^\top$ 
and $\forall \ell$, for which we employ an approximation based on finite $M$ constraints \citep{monotonicGPR}.
Define 
$\*U^{\prime} = \left(\pd{U(\*f^{N+1})}{f^{(N+1)}_{\ell_1}},\ldots, \pd{U(\*f^{(N+M)})}{f^{(N+M)}_{\ell_M}}\right)^{\top}$, 
where 
$\*f^{(N+1)}, \ldots, \*f^{(N+M)}$
and 
$\ell_1, \ldots, \ell_M$
are the points and their dimensions on which the derivative constraints are imposed (see \citep{monotonicGPR} for the selection of these points and dimensions).
The prior covariance of the joint density 
$p(\*U, \*U^{\prime}) = \mathcal{N}(\*0, \*K_{\rm{joint}})$
becomes
\begin{align*}
\left(\*{K}_{\text {joint }}\right)_{i j}= 
 \begin{cases}
 k\left(\*f^{(i)}, \*f^{(j)}\right) & \text { for } i=1, \ldots, N, j=1, \ldots, N \\ 
  \frac{\partial}{\partial f_{\ell_{i-N}}^{(i)}} k\left(\*f^{(i)}, \*f^{(j)}\right) & \text { for } i=N+1, \ldots, N+M, j=1, \ldots, N \\ 
  \left(\*{K}_{\text {joint }}\right)_{j i} & \text { for } i=1, \ldots, N, j=N+1, \ldots, N+M \\ 
  \frac{\partial^2}{\partial f_{\ell_{i-N}}^{(i)} \partial f_{\ell_{j-N}}^{(j)}} k\left(\*f^{(i)}, \*f^{(j)} \right) & \text { for } i=N+1, \ldots, N+M, j=N+1, \ldots, N+M.
 \end{cases}
\end{align*}

To incorporate monotonicity information, \citet{monotonicGPR} replace the non-negativity constraint on the derivative with the likelihood of a variable $m^{(i)}_{\ell_i}$ that represents the derivative information at the dimension $\ell_i$ of $\*f^{(N+i)}$:
\begin{align*}
 p\left( m^{(i)}_{\ell_i} \mid 
 \pd{U(\*f^{(N+i)})}{f^{(N+i)}_{\ell_i}} 
 \right) = 
 \Phi\left(\pd{U(\*f^{(N+i)})}{f^{(N+i)}_{\ell_i}} \frac{1}{\nu}\right),
\end{align*}
where the hyper-parameter $\nu$ controls the strictness of monotonicity information (set as $\nu=10^{-6}$ in our experiments).
%
Using the preference information \eq{eq:likelihood-PC} and derivative information \citep{monotonicGPR}, the posterior distribution of monotonic preference GP is
\begin{align}
p(\*U, \*U^{\prime} | D_{\rm{pre}}, \*m) &\propto p(\*U, \*U^{\prime}) p(D_{\rm{pre}} | \*U) p(\*m | \*U^{\prime})\notag\\
&= \mathcal{N}(\*0, \*K_{\rm{joint}})\prod^n_{j=1} \Phi\left(\frac{U(\*f^{(j)}) - U(\*f^{(j^\prime)})}{\sqrt{ 2 } \sigma_{\rm PC} } \right) \prod^M_{i=1} \Phi\left(\pd{U(\*f^{(N+i)})}{f_{\ell_i}^{(N+i)}}\frac{1}{\nu}\right),
\label{eq:pregp-posterior}
\end{align}
where $D_{\rm{pre}}$ is PC observations and 
$\*m=\left(m_{\ell_1}^{(1)},\ldots, m_{\ell_M}^{(M)}\right)^{\top}$.

The analytical calculation of \eq{eq:pregp-posterior} is difficult and we employ the Expectation Propagation (EP) \citep{minka2001expectation}.
See Appendix~\ref{app:pregpmono} for the detail of EP.
EP provides an approximate joint posterior of $\*U$ and $\*U^\prime$ as a multi-variate Gaussian distribution, denoted as $p(\*U,\*U^\prime \mid D_{\rm pre}, \*m) \approx \cN(\*\mu, \*\Sigma)$, where $\*\mu$ and $\*\Sigma$ are the approximate posterior mean and covariance matrix.
As a result, we obtain the predictive distribution $\cN(\mu(\*f_*), \sigma^2(\*f_*))$ for any $\*f_* \in \RR^L$ as 
\begin{align*}
\mu(\*f_{*}) &= \*k_{*\rm joint}^{\top}\*\Sigma^{-1}\*\mu,\\
\sigma^2(\*f_{*}) &= k(\*f_{*}, \*f_{*}) - \*k_{*\rm joint}^{\top}\*\Sigma^{-1}\*k_{*\rm joint},
\end{align*}
where the $i$-th element of $\*k_{*\rm joint}$ is
\begin{align*}
(k_{*\rm joint})_i = 
\begin{cases}
k(\*f_{*}, \*f^{(i)}) & (i = 1,\ldots, N),\\
\pd{}{f_{\ell_{i-N}}^{(i)}} k(\*f_{*}, \*f^{(i)}) & (i = N+1,\ldots, N+M).
\end{cases}
\end{align*}

However, because of its high flexibility, the GP model may require a larger number of observations to estimate the DM preference accurately. 
In this sense, the simple CSF-based and the GP-based approaches are expected to have trade-off relation about their flexibility and complexity of the estimation.

\section{Related Work}
\label{s:related-work}

In the MOO literature, incorporating the DM preferences into exploration algorithms have been studied.
For example, a hyper-rectangle or a vector are often used to represent the preference of the DM in the objective function space \citep[e.g.,][]{hakanen2017using, palar2018multi, he2020preference, mobors}.
Another example is that \citet{abdolshah2019multi} represent the DM preference by the order of importance for the objective functions.
Furthermore, \citet{ahmadianshalchi2023preference} incorporate importance for the objective functions based on a vector given by the DM.
In particular, \citet{mobors} use a similar formulation to our approach in which CSF with the random parameters can be used for the utility function.
These approaches do not estimate a model of the DM preference, by which the DM needs to directly specify the detailed requirement of the performance though it is often difficult in practice.
For example, in the case of the hyper-rectangle, there may not exist any solutions in the specified region by the DM. 


The interactive preference estimation with MOO has also been studied mainly in the context of evolutionary algorithms \citep{hakanen2016connections}.
For example, \citet{taylor2021bayesian} combine preference learning with the multi-objective evolutionary algorithm.
On the other hand, to our knowledge, combining an interactively estimated Bayesian preference model and the multi-objective BO \citep{emmerich2008computation,mobors,belakaria2019max,hernandez2016predictive,suzuki2020multi,knowles2006parego} has not been studied, though multi-objective extension of BO has been widely studied \citep[e.g.,][]{Emmerich2005-Single,Hernandez2014-Predictive}. 
\citet{multi-att} and \citet{lin2022preference} consider a different preference based BO in which the DM can have an arbitrary preference over the multi-dimensional output space, meaning that the problem setting is not MOO anymore (in the case of MOO, the preference should be monotonically non-decreasing). 
Further, these studies do not discuss the IR-type supervision.
%
%

Instead of separately estimating $\*f(\*x)$ and $U(\*f)$, it is also possible that directly approximating the DM preference $U(\*f(\*x))$ as a function of $\*x$, denoted as $\tilde{U}(\*x) = U(\*f(\*x))$, through some preferential model such as the preferential GP.
For maximizing $\tilde{U}(\*x)$, preferential BO \citep{eric2007active, brochu2010interactive, gonzalez2017preferential, takeno2023towards} can be used.
However, in this approach, the DM preference can be queried only about already observed $\*f(\*x)$.
For example, in the PC observation between $\*x$ and $\*x^\prime$, the DM can provide their relative preference only after $\*f(\*x)$ and $\*f(\*x^\prime)$ are observed.
On the other hand, in our proposed framework, the preference model $U(\*f)$ is defined on the space of the objective function values $\*f \in \RR^L$, by which preference of any pair of $\*f$ and $\*f^\prime$ can be queried to the DM anytime.
Further, this $\tilde{U}(\*x)$ based approach ignores the existence of $\*f$ in the surrogate model, by which the information of the observed values for $\*f(\*x)$ is not fully used and the monotonicity constraint is also not incorporated.

\begin{figure}
 \begin{center}
  \subfloat[$L = 2$]{\igr{.35}{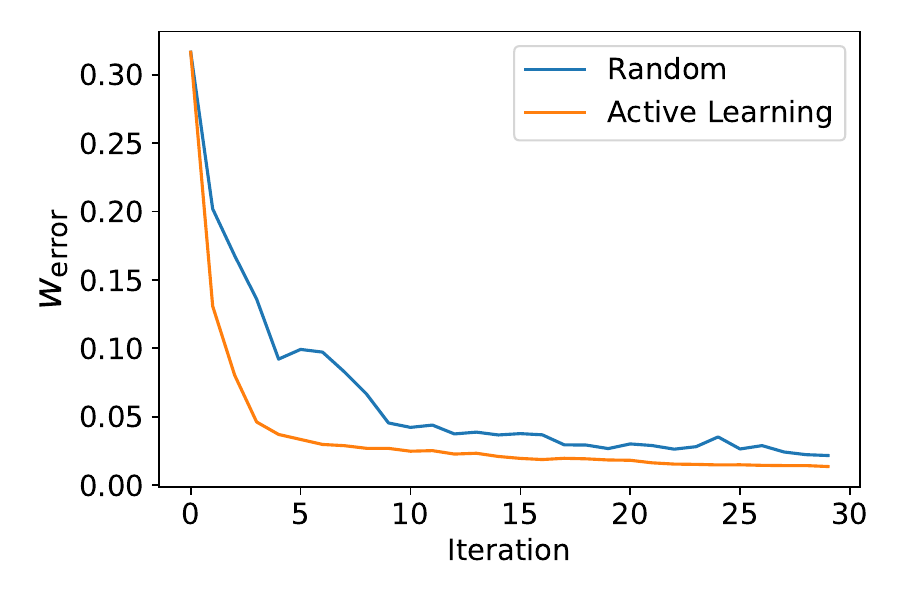}}
  \subfloat[$L = 10$]{\igr{.35}{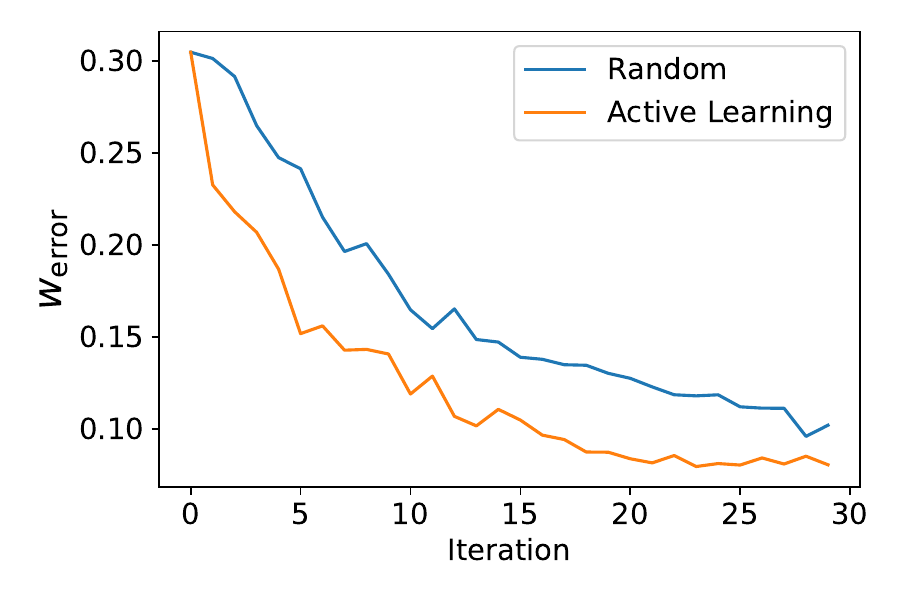}}
 \end{center}
 \caption{
 Estimation error of $\*w$. 
 }
 \label{fig:active-learning-ex}
\end{figure}
\begin{figure}
 \begin{center}
  \igr{.35}{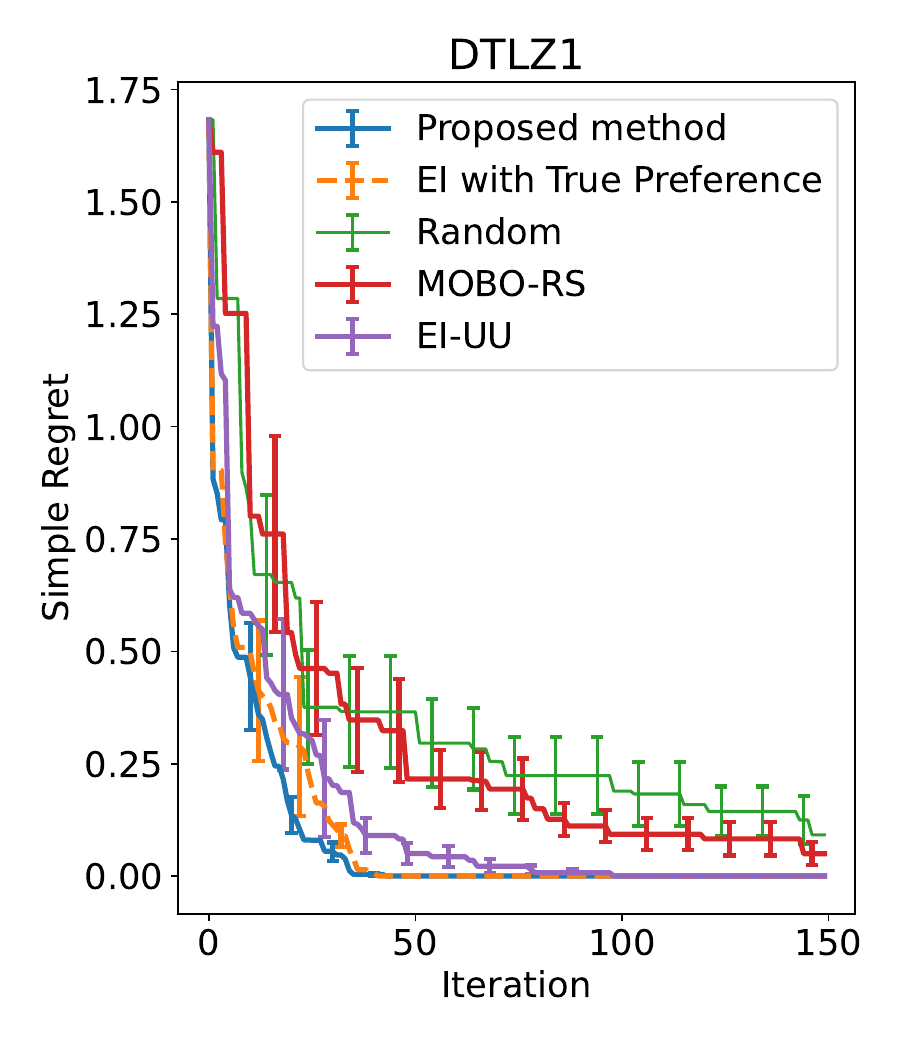}%
  \nobreak
  \igr{.35}{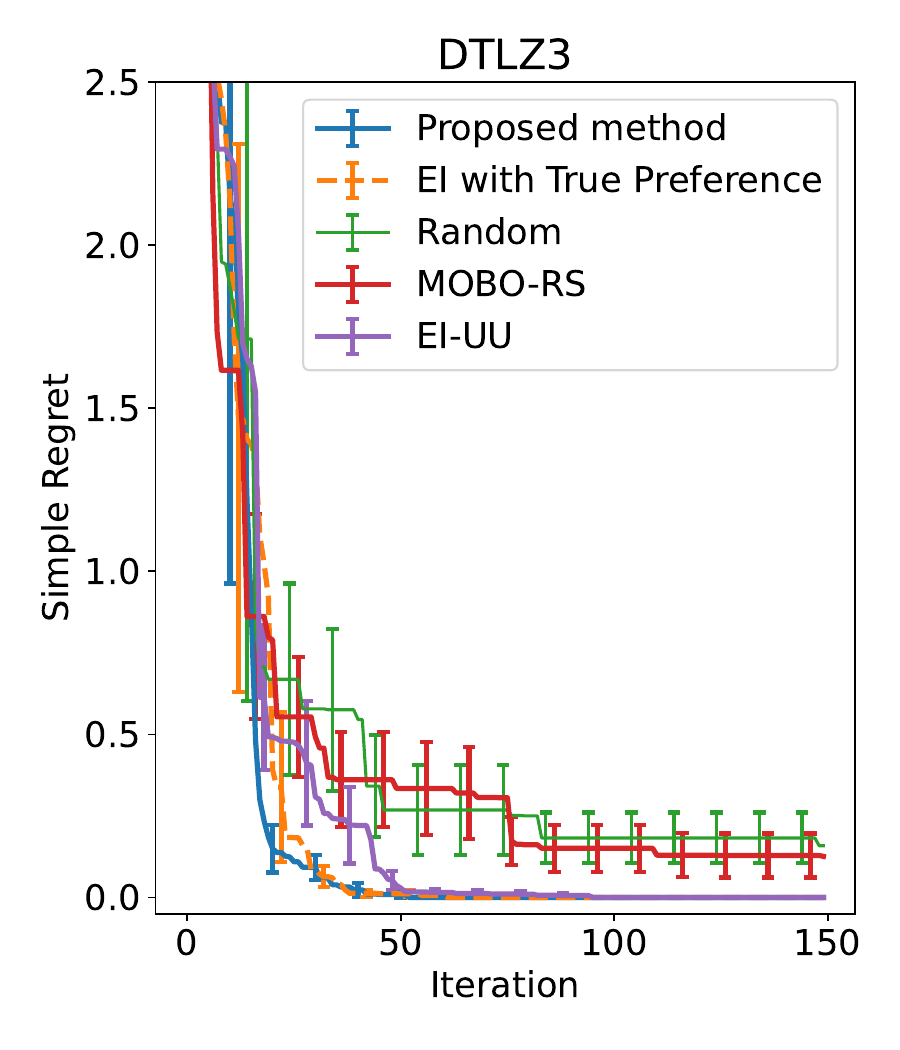}

  \igr{.35}{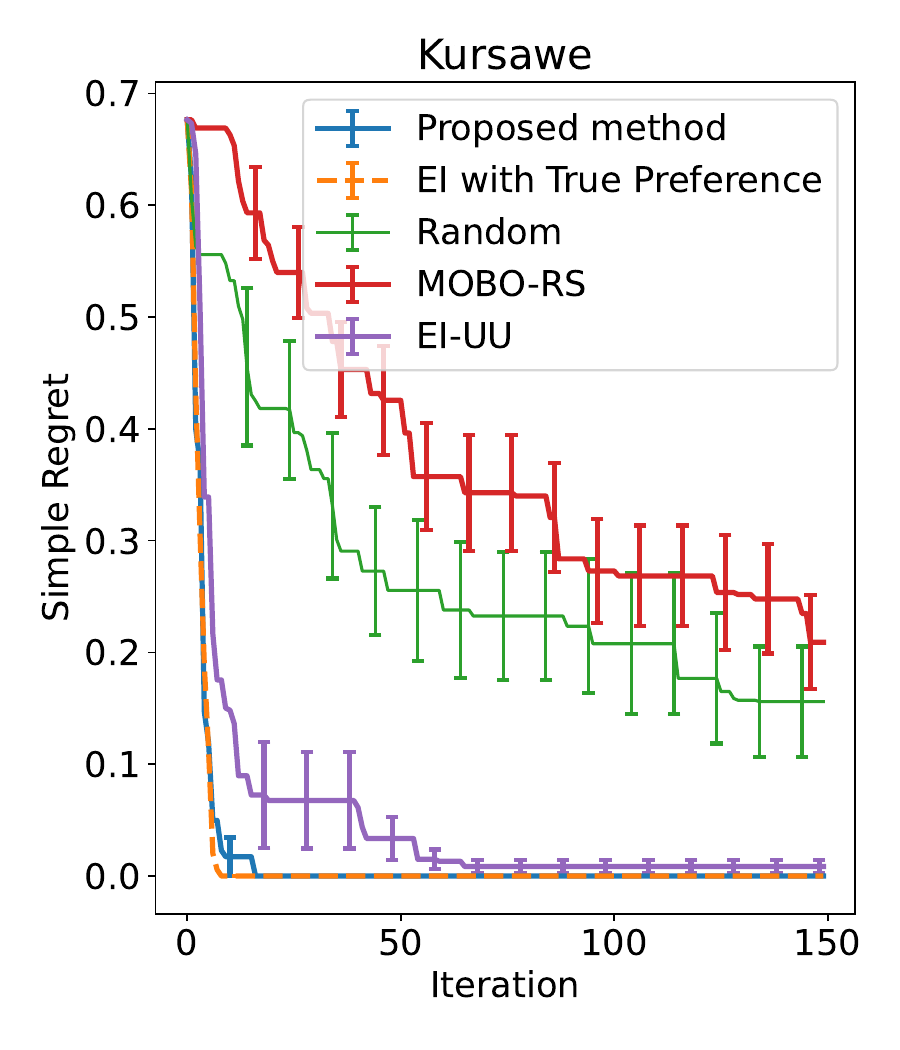}\nobreak
  \igr{.35}{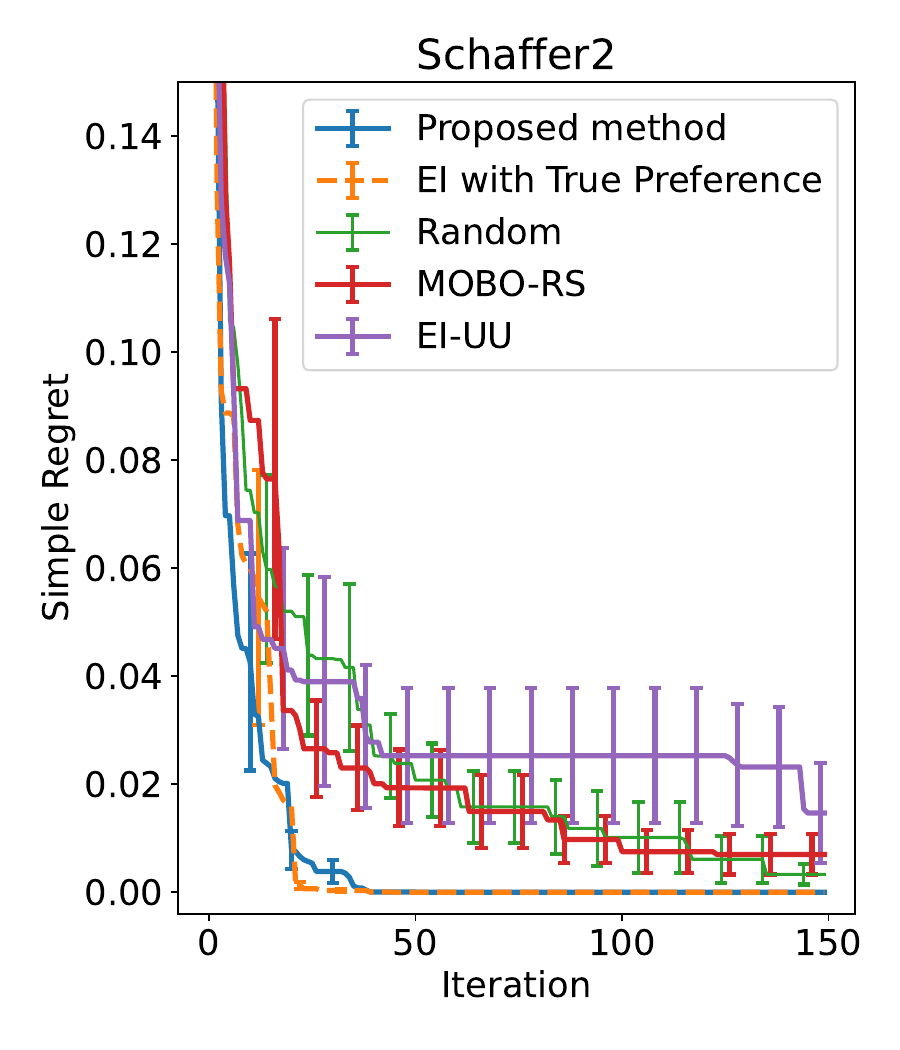}
 \end{center}
 \caption{
 Simple regret on benchmark functions.
 }
 \label{fig:bo-benchmark-ex}
\end{figure}
 
\section{Experiments}
\label{s:experiments}

We perform two types of experiments.
First, in \S~\ref{ss:experiment-PL}, we evaluate the performance of our MI-based active learning (described in \S~\ref{sss:acqs}) that efficiently learns the preference model \eq{eq:Cheby}.
Next, in \S~\ref{ss:experiment-BO}, we evaluate the performance of the entire framework of our proposed method, for which we used a benchmark function and two settings of hyper-parameter optimization of machine learning models.
In all experiments (excluding the experiment with nonparametric utility function in \S~\ref{ss:experiment-pregp}), the true utility function (the underlying true DM preference) is defined by \eqref{eq:Cheby} with the parameter $\*w_{\rm true}$, determined through the sampling from the Dirichlet distribution ($\*\alpha = (2, \ldots, 2)^\top$).
%
%
%
GPs for $\*f(\*x)$ employs the RBF kernel.
Preference observations are generated with the noise variance 
$\sigma_{\rm PC} = \sigma_{\rm IR} = 0.1$.
%
See Appendix~\ref{app:setting-bo} for other settings of BO.
%

\subsection{Preference Learning with Active Query Selection}
\label{ss:experiment-PL}

We evaluate estimation accuracy of the preference parameter $\*w$.
Our MI-based acquisition function and the random query selection were compared.
We iteratively added the preference information of the DM.
At every iteration, a PC observation and an IR observation are provided. 
%

\figurename~\ref{fig:active-learning-ex} shows the results.
In each plot, the horizontal axis is the iteration and the vertical axis is 
$w_{\rm error} = \frac{1}{T}\sum_{t=1}^{T}\|\*w_{\text{true}} - \*w_t\|_2$, 
where $\*w_1, \ldots, \*w_T$ are sampled from the posterior ($T = 1000$). 
The results are the average of 10 runs.
We can see that the error rapidly decreases, and further, active learning obviously improves the accuracy.
Even when $L = 10$, $w_{\rm error}$ decreased to around 0.1 only with a few tens of iterations. 
Since $w_{\rm error}$ becomes 0 only when the posterior generates the exact $w_{\rm true}$ $T$ times, we consider $w_{\rm error} \approx 0.1$ is sufficiently small to accelerate our preference based BO (note that $\|\*w_t \|_1 = 1$). 

\subsection{Utility Function Optimization}
\label{ss:experiment-BO}

We evaluate efficiency of the proposed method by evaluating the simple regret on the true utility function $U_{\*w_{\rm true}}$:
\begin{align*}
 \max_{\*x} 
 U_{\*w_{\rm true}}(\*f(\*x))
 - \max_{\*x \in \Theta_s} U_{\*w_{\rm true}}(\*f(\*x)), 
\end{align*}
where $\Theta_s$ is a set of $\*x$ already observed.
This evaluates the difference of the utility function values between the true optimal (the first term) and the current best value (the second term).
The true parameter $\*w_{\rm true}$ was determined through the sampling from the Dirichlet distribution.
We assume that a PC- and an IR- observation are obtained when the DM provides the preference information. 
The results are shown in the average and the standard error of $10$ runs.
The performance was compared with the three methods: random selection (Random), a random scalarization-based multi-objective BO, called MOBO-RS \citep{mobors}, and multi-attribute BO with preference-based expected improvement acquisition function, called EI-UU \citep{multi-att}. 
%
Further, we evaluate the performance of EI \eq{eq:EI} with the fixed ground-truth preference vector $\*w_{\rm true}$, referred to as EI with True Preference (EI-TP).
We interpret EI-TP as a best possible baseline for our proposed method, because it has a complete DM preference from the beginning.

We used benchmark functions and two problem settings of hyper-parameter optimization for machine learning models (cost-sensitive learning and fairness-aware machine learning).

\subsubsection{Benchmark Function}
\label{ss:experiment-benchmark}

We use well-known MOO benchmark functions, called DTLZ1 and DTLZ3 \citep{deb2005scalable}.
In both the functions, the input and output dimensions are $d = 3$ and $L = 3$.
In addition, we use benchmark functions called 
Kursawe \citep{kursawe1990variant}, 
and Schaffer2 \citep{schaffer1985multiple}, in which 
$(d,L) = (3,2)$ and $(d,L) = (1,2)$, 
respectively.
%
Detailed settings of these functions are in Appendix~\ref{app:setting-bench}.
%
We prepare 1000 input candidates by taking grid points in each input dimension.
At every iteration, a PC- and an IR- observation are selected by active learning, and the posterior of $\*w$ is updated.

The results are shown in \figurename~\ref{fig:bo-benchmark-ex}.
%
We see that our proposed method can efficiently reduce simple regret compared with Random, MOBO-RS (which seeks the entire Pareto front), and EI-UU. 
Note that since EI-UU only incorporates PC observations (does not use IR), we show the comparison between the proposed method only with PC (without IR) and EI-UU in Appendix~\ref{app:comparison-pc-uu}.
%
On the other hand, the proposed method and EI-TP show similar performance, which indicates that the posterior of $\*w$ provides sufficiently useful information for the efficient exploration. 
%

%

We further provide results on other benchmark functions (Appendix~\ref{app:additional-bench}), sensitivity evaluation to the number of samplings in EI (Appendix~\ref{app:sensitivity}), and ablation study on PC and IR (Appendix~\ref{app:ablation}).

\begin{figure}
 \begin{center}
\igr{.35}{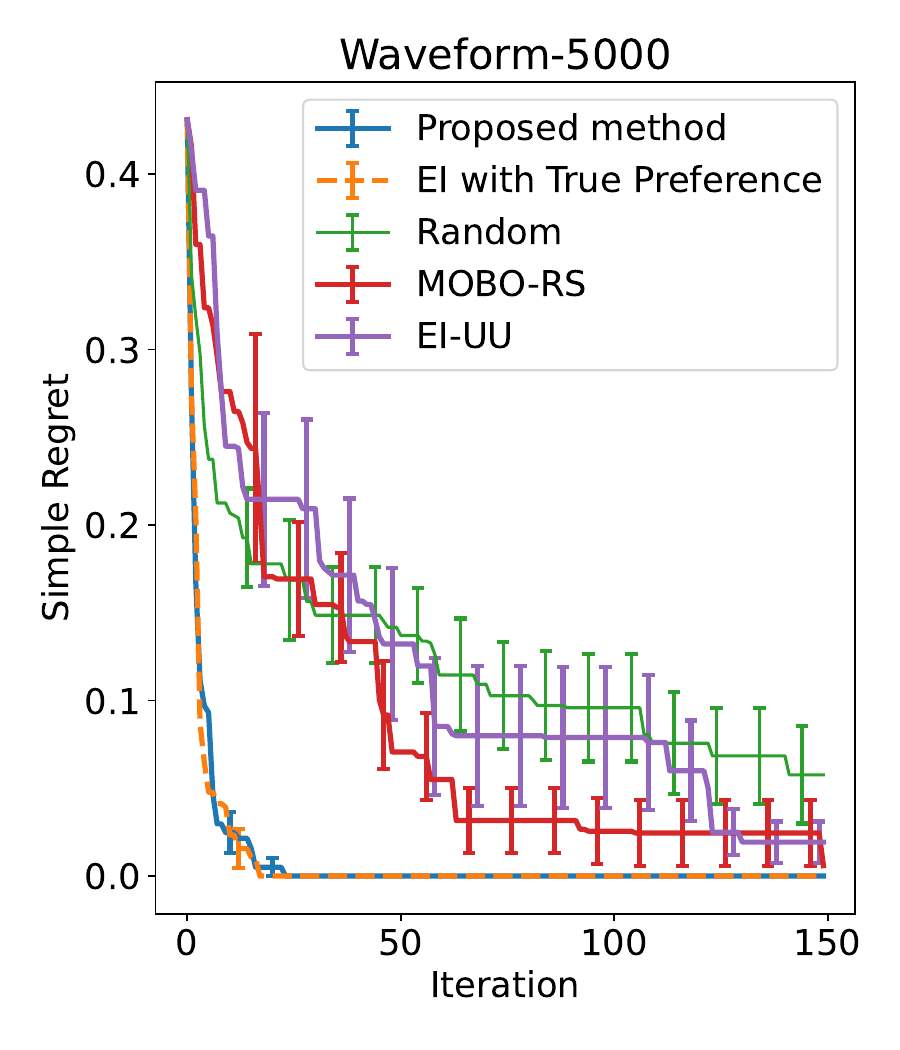}\nobreak
\igr{.35}{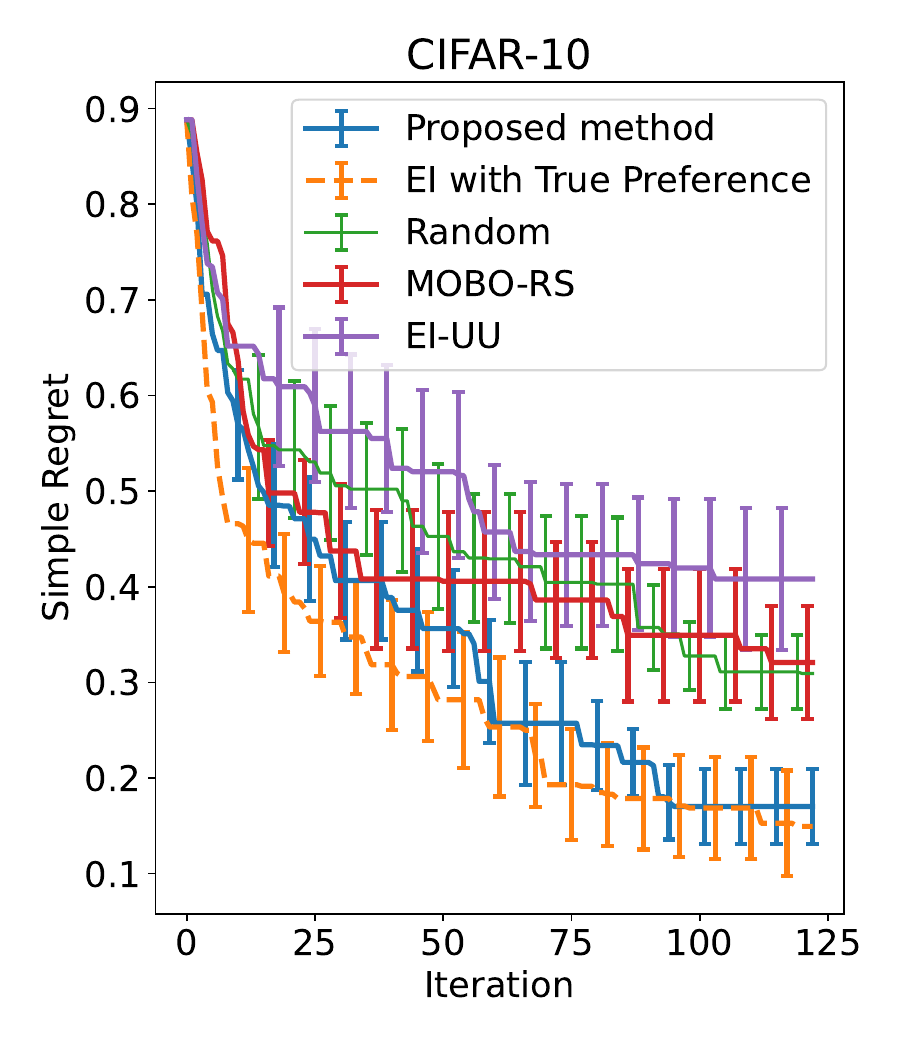}
 \end{center}
 \caption{
 Simple regret on the hyper-parameter optimization for cost-sensitive learning.
 }
 \label{fig:bo-classweight-ex}
 \begin{center}
\igr{.35}{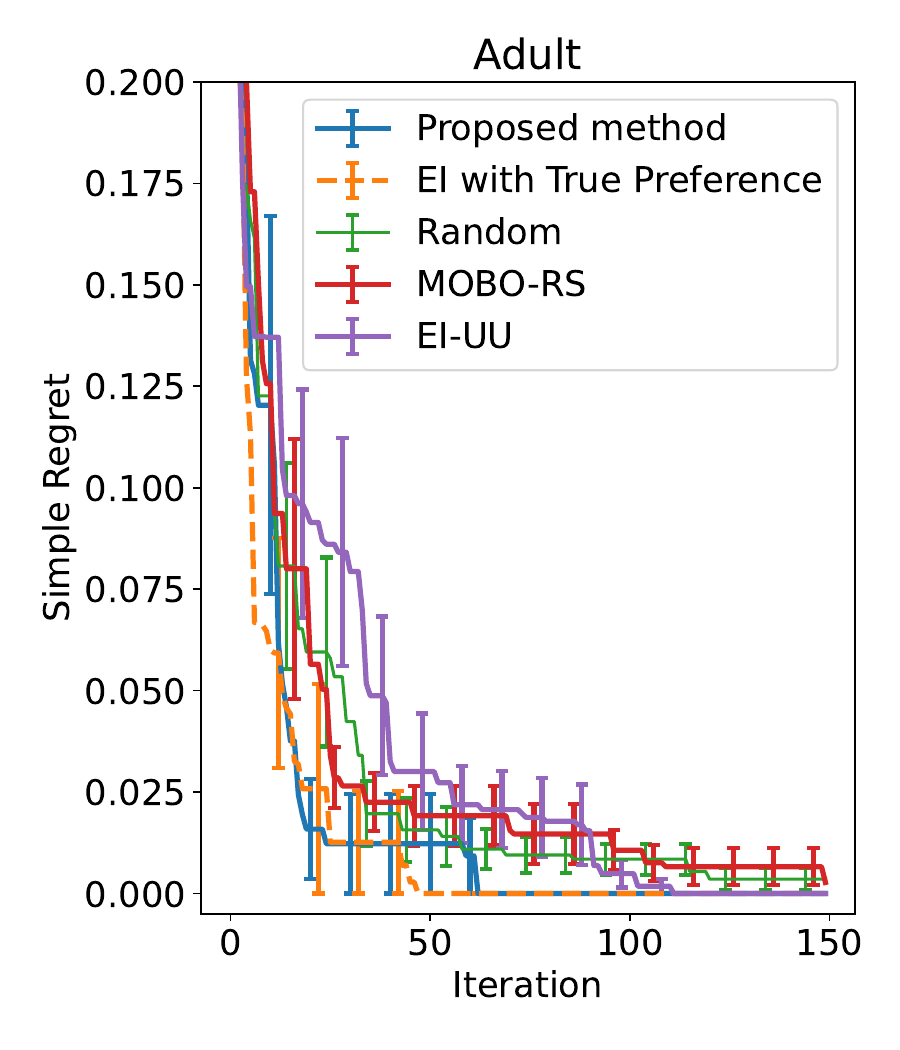}\nobreak
\igr{.35}{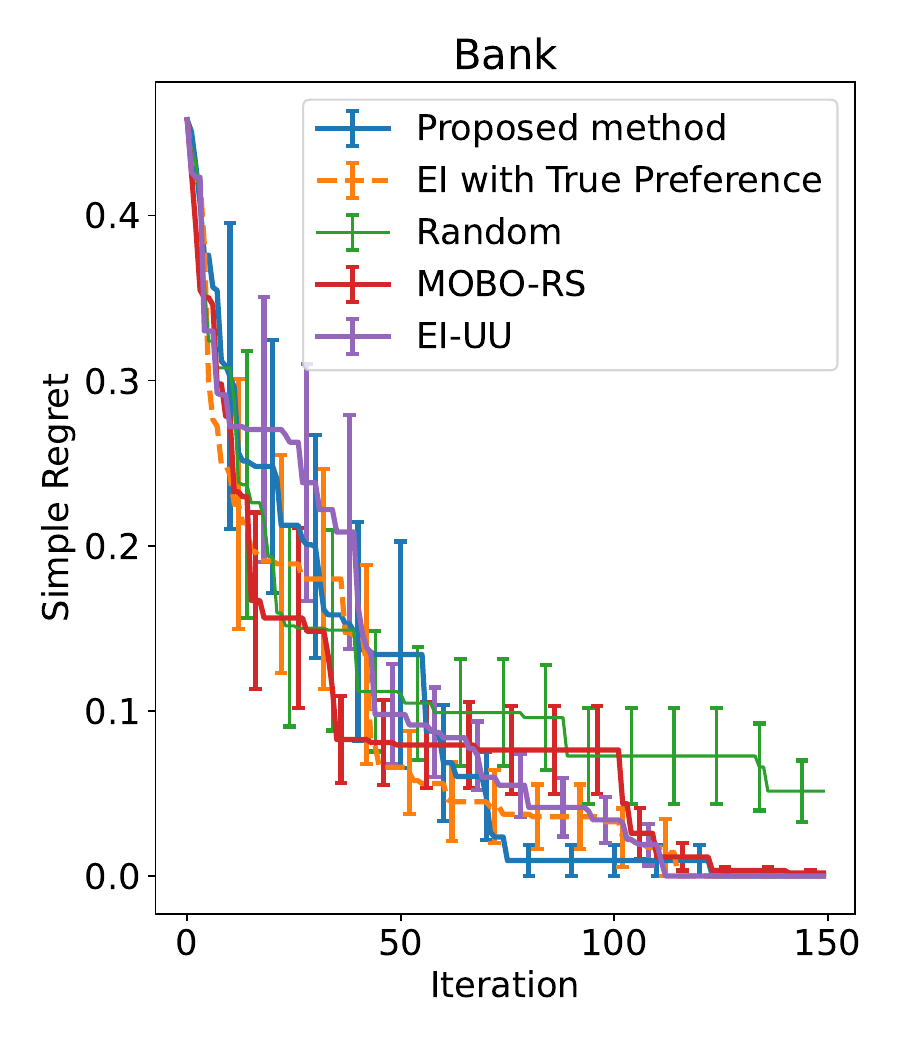}
 \end{center}
 \caption{
 Simple regret on the hyper-parameter optimization for fairness-aware learning.
 }
 \label{fig:bo-fairness-ex}
\end{figure}



\subsubsection{Hyper-parameter Optimization for Cost-sensitive Learning}
\label{ss:experiment-HPO}

In many real-world applications of multi-class classification, a DM may have different preferences over each type of misclassifications \citep{elkan2001foundations, autobalance}.
%
For example, in a disease diagnosis, the miss-classification cost for the disease class can be higher than those of the healthy class.
In learning algorithms, the cost balance are often controlled by hyper-parameters.
In this section, we consider a hyper-parameter optimization problem for $L$-class classification models considering the importance of each class.

For the hyper-parameter optimization of cost-sensitive learning, we used LightGBM \citep{ke2017lightgbm} and neural networks.
For LightGBM, $\*x$ is defined by the $L$-dimensional `class\_weight' parameters, which controls the importance of each class. 
In the case of neural networks, the weighted cross-entropy loss
$\sum_{i=1}^L - \lambda_i y_i \log \hat{y}_i$ was used, 
where $\lambda_i > 0$ is the class weight, $y_i \in \{0,1\}$ is one-hot encoding of the label and $\hat{y}_i$ is the corresponding class predicted probability. 
In this case, $\*x$ consists of  
$\lambda_1, \ldots, \lambda_L$.
The objective functions $\*f(\*x)$ are defined by recall of each class on the validation set (e.g. $f_1$ represents recall of the first class, $f_2$ represents recall of the second class, ...).
%
%
%
The datasets for the classifiers are Waveform-5000 ($L = 3$) and CIFAR-10 ($L = 10$) \citep{cifar10}, for which details are in Appendix~\ref{app:setting-hpo}. 
For Waveform-5000, we used LightGBM.
For CIFAR-10, we used Resnet18 \citep{resnet} pre-trained by Imagenet \citep{deng2009imagenet}.
%
%
%
The input dimension (dimension of hyper-parameters) equals to the number of classes, i.e., $d = L$.

\figurename~\ref{fig:bo-classweight-ex} shows the results.
Overall, we see that the same tendency with the case of benchmark function optimization.
The proposed method outperformed MOBO-RS and EI-UU, and was comparable with EI-TP.
%
In the CIFIR-10 dataset, which has the highest output dimension ($10$), EI-TP was better than the proposed method.
When the output dimension is high, the preference estimation can be more difficult, but we still obviously see that the proposed method drastically accelerates the exploration compared with searching the entire Pareto front.

In Appendix~\ref{app:additional-hpo}, we show additional results on this problem setting.

\subsubsection{Hyper-parameter Optimization for Fair Classification}
\label{ss:experiment-fairness}

As another example of preference-aware hyper-parameter optimization, we consider the fairness in classification problem.
%
Although specific sensitive attributes (e.g. gender, race) should not affect the prediction results for fairness, generally high fairness degrades the classification accuracy \citep{zafar2017fairness}.
%
%
\citet{zafar2017fairness} propose the classification method which maximizes fairness under accuracy constraints, and it has a hyper-parameter named $\gamma$ to control trade-off between fairness and accuracy.

We consider the hyper-parameter optimization with $2$-dimensional objective functions consisting of the level of fairness (p\%-rule \citep{biddle2006adverse}) and accuracy of the classification.
%
We use the logistic regression classifier proposed in \citep{zafar2017fairness}, and the hyper-parameter $\gamma$, which controls trade-off between fairness and accuracy, is seen as the input $\*x$ ($d = 1$).
For two datasets, Adult \citep{misc_adult_2} and Bank \citep{misc_bank_marketing_222}, we prepare $201$ candidates of $\*x$ by taking grid points in $[0, 1]$.
%
The explanatory variable ``sex'' in the Adult dataset and ``age'' in the Bank dataset are regarded as sensitive attributes, respectively, and their fairness is considered in the logistic regression classifier.
Other settings including the data preprocessing comply with \citet{zafar2017fairness}.

\figurename~\ref{fig:bo-fairness-ex} shows the results.
%
In comparison to other methods, the proposed method exhibited a rapid decrease in the simple regret and identified the best solution in fewer iterations, particularly for the Adult dataset.

\subsubsection{Evaluation with GP-based Utility Function}
\label{ss:experiment-pregp}

\begin{figure}
 \begin{center}
  \igr{.4}{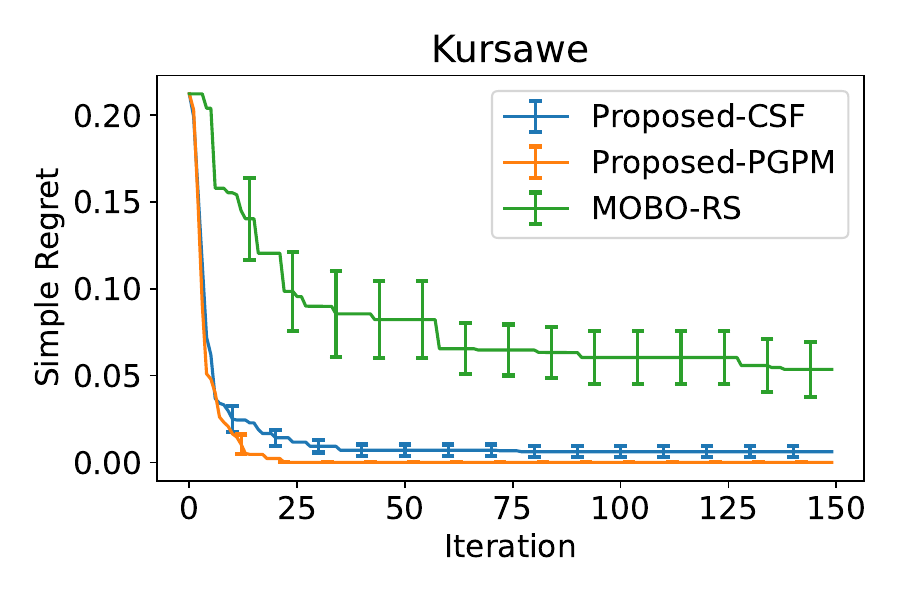}
  \igr{.4}{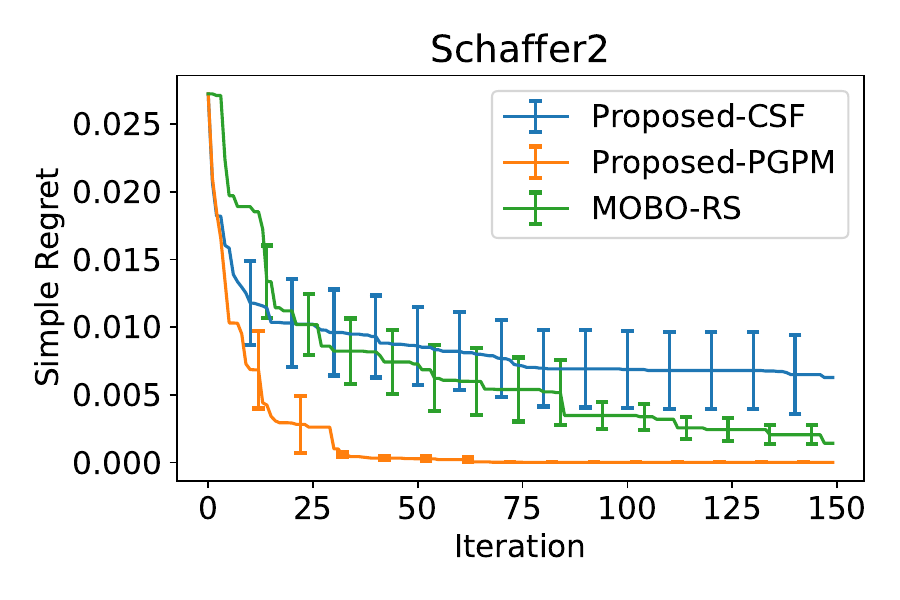}
 \end{center}
 \caption{Simple regret on benchmark functions using the CSF model and the GP-based model.}
 \label{fig:preferential-gp}
\end{figure}

%
%
%
We perform experiments with more flexible utility functions.
%
%
We set the true utility function using the following basis function model:
\begin{align*}
U_{\rm true}(\*f) &= \sum_{i=1}^M \lambda_i \phi_i(\*f), 
\end{align*}
where 
\begin{align*}
\phi_i(\*f) &= \frac{1}{1 + \exp(-(\*\beta_i^{\top}\*f + \beta_i))}, 
\end{align*}
and this model has parameters
$\*\beta_i \in \RR^L$, 
$\beta_i \in \RR$,
and
$\lambda_1, \ldots, \lambda_M$.
%
All the elements in the coefficients
$\lambda_1, \ldots, \lambda_M$
and the parameter vector 
$\*\beta_i$
are non-negative values that makes $U_{\rm true}(\*f)$ a monotonically non-decreasing function.
%
We randomly generate $\beta_i$ and each element of $\*\beta_i$ from the uniform distribution in $[0,1]$, and generate $\lambda_i$ from the normal distribution $N(0, 1)$ truncated so that its support is $[0,\infty)$ (i.e., truncated normal distribution). 

We apply our proposed method with two different utility functions: the CSF-based utility function \eq{eq:Cheby} and the preferential GP with monotonicity information, called PGPM. 
We use the benchmark functions Kursawe and Schaffer2, already shown in \S~\ref{ss:experiment-benchmark}. 
The results are in \figurename~\ref{fig:preferential-gp}.
Here, we only use PC as a weak supervision for the utility function estimation.
%
In \figurename~\ref{fig:preferential-gp}, Proposed-CSF and 
Proposed-PGPM
mean the proposed method using the CSF and PGPM, respectively.
%
We see that Proposed-CSF does not perform better than Proposed-PGPM.
%
This indicates that, for a complicated true utility function, PGPM can be effective than the simple CSF.
Additional results with other utility functions and benchmark functions are shown in Appendix~\ref{app:exp-other-utility}.
%

\section{Conclusion}
\label{s:conclusion}

We proposed a multi-objective Bayesian optimization (BO) method in which the preference of the decision maker (DM) is adaptively estimated through a human-in-the-loop manner.
The DM's preference is represented by a Chebyshev scalarization based utility function, for which we assume that pairwise comparison (PC) and improvement request (IR) are provided as weak supervisions by a decision maker (DM).
Our acquisition function is based on the well-known expected improvement by which uncertainty of both the original objective functions and the preference model can be incorporated.
We further proposed a mutual information based active learning strategy that reduces the interaction cost with the DM.
Empirical evaluation indicated that our proposed method accelerates the optimization on several benchmark functions, and applications to hyper-parameter optimization of machine learning models are also shown.


\section*{Acknowledgements}
This work was supported by MEXT KAKENHI (20H00601, 21H03498, 22H00300, 23K17817), JST CREST (JPMJCR21D3, JPMJCR22N2), JST Moonshot R\&D (JPMJMS2033-05), JST AIP Acceleration Research (JPMJCR21U2), NEDO (JPNP18002, JPNP20006), RIKEN Center for Advanced Intelligence Project, JSPS KAKENHI Grant Number JP21J14673, and JST ACT-X (JPMJAX23CD).

\bibliography{ref}

\begin{thebibliography}{48}
\providecommand{\natexlab}[1]{#1}
\providecommand{\url}[1]{\texttt{#1}}
\expandafter\ifx\csname urlstyle\endcsname\relax
  \providecommand{\doi}[1]{doi: #1}\else
  \providecommand{\doi}{doi: \begingroup \urlstyle{rm}\Url}\fi

\bibitem[Abdolshah et~al.(2019)Abdolshah, Shilton, Rana, Gupta, and
  Venkatesh]{abdolshah2019multi}
M.~Abdolshah, A.~Shilton, S.~Rana, S.~Gupta, and S.~Venkatesh.
\newblock Multi-objective bayesian optimisation with preferences over
  objectives.
\newblock In H.~Wallach, H.~Larochelle, A.~Beygelzimer, F.~d\textquotesingle
  Alch\'{e}-Buc, E.~Fox, and R.~Garnett, editors, \emph{Advances in Neural
  Information Processing Systems}, volume~32. Curran Associates, Inc., 2019.

\bibitem[Ahmadianshalchi et~al.(2023)Ahmadianshalchi, Belakaria, and
  Doppa]{ahmadianshalchi2023preference}
A.~Ahmadianshalchi, S.~Belakaria, and J.~R. Doppa.
\newblock Preference-aware constrained multi-objective bayesian optimization.
\newblock \emph{arXiv preprint arXiv:2303.13034}, 2023.

\bibitem[Alvarez et~al.(2012)Alvarez, Rosasco, Lawrence, et~al.]{kernel}
M.~A. Alvarez, L.~Rosasco, N.~D. Lawrence, et~al.
\newblock Kernels for vector-valued functions: A review.
\newblock \emph{Foundations and Trends{\textregistered} in Machine Learning},
  4\penalty0 (3):\penalty0 195--266, 2012.

\bibitem[Astudillo and Frazier(2020)]{multi-att}
R.~Astudillo and P.~Frazier.
\newblock Multi-attribute bayesian optimization with interactive preference
  learning.
\newblock In \emph{International Conference on Artificial Intelligence and
  Statistics}, pages 4496--4507. PMLR, 2020.

\bibitem[Bechikh et~al.(2015)Bechikh, Kessentini, Said, and
  Gh\UTF{00E9}dira]{bechikh2015preference}
S.~Bechikh, M.~Kessentini, L.~B. Said, and K.~Gh\UTF{00E9}dira.
\newblock Preference incorporation in evolutionary multiobjective optimization:
  A survey of the state-of-the-art.
\newblock volume~98 of \emph{Advances in Computers}, pages 141--207. Elsevier,
  2015.

\bibitem[Becker and Kohavi(1996)]{misc_adult_2}
B.~Becker and R.~Kohavi.
\newblock {Adult}.
\newblock UCI Machine Learning Repository, 1996.
\newblock {DOI}: https://doi.org/10.24432/C5XW20.

\bibitem[Belakaria et~al.(2019)Belakaria, Deshwal, and Doppa]{belakaria2019max}
S.~Belakaria, A.~Deshwal, and J.~R. Doppa.
\newblock Max-value entropy search for multi-objective bayesian optimization.
\newblock \emph{Advances in neural information processing systems}, 32, 2019.

\bibitem[Biddle(2006)]{biddle2006adverse}
D.~Biddle.
\newblock \emph{Adverse impact and test validation: A practitioner's guide to
  valid and defensible employment testing}.
\newblock Gower Publishing, Ltd., 2006.

\bibitem[Brochu(2010)]{brochu2010interactive}
E.~Brochu.
\newblock \emph{Interactive Bayesian optimization: learning user preferences
  for graphics and animation}.
\newblock PhD thesis, University of British Columbia, 2010.

\bibitem[Chu and Ghahramani(2005{\natexlab{a}})]{chu2005extensions}
W.~Chu and Z.~Ghahramani.
\newblock Extensions of gaussian processes for ranking: semisupervised and
  active learning.
\newblock \emph{Learning to Rank}, 29, 2005{\natexlab{a}}.

\bibitem[Chu and Ghahramani(2005{\natexlab{b}})]{pregp}
W.~Chu and Z.~Ghahramani.
\newblock Preference learning with {G}aussian processes.
\newblock In \emph{Proceedings of the 22nd international conference on Machine
  learning}, pages 137--144, 2005{\natexlab{b}}.

\bibitem[Deb et~al.(2005)Deb, Thiele, Laumanns, and Zitzler]{deb2005scalable}
K.~Deb, L.~Thiele, M.~Laumanns, and E.~Zitzler.
\newblock \emph{Scalable test problems for evolutionary multiobjective
  optimization}.
\newblock Springer, 2005.

\bibitem[Deng et~al.(2009)Deng, Dong, Socher, Li, Li, and
  Fei-Fei]{deng2009imagenet}
J.~Deng, W.~Dong, R.~Socher, L.-J. Li, K.~Li, and L.~Fei-Fei.
\newblock Imagenet: A large-scale hierarchical image database.
\newblock In \emph{2009 IEEE conference on computer vision and pattern
  recognition}, pages 248--255. Ieee, 2009.

\bibitem[Dua and Graff(2017)]{Dua:2019}
D.~Dua and C.~Graff.
\newblock {UCI} machine learning repository, 2017.
\newblock URL \url{http://archive.ics.uci.edu/ml}.

\bibitem[Elkan(2001)]{elkan2001foundations}
C.~Elkan.
\newblock The foundations of cost-sensitive learning.
\newblock In \emph{Proceedings of the 17th International Joint Conference on
  Artificial Intelligence - Volume 2}, IJCAI'01, page 973\UTF{2013}978, San
  Francisco, CA, USA, 2001. Morgan Kaufmann Publishers Inc.

\bibitem[Emmerich and Klinkenberg(2008)]{emmerich2008computation}
M.~Emmerich and J.-w. Klinkenberg.
\newblock The computation of the expected improvement in dominated hypervolume
  of pareto front approximations.
\newblock \emph{Rapport technique, Leiden University}, 34:\penalty0 7--3, 2008.

\bibitem[Emmerich(2005)]{Emmerich2005-Single}
M.~T.~M. Emmerich.
\newblock Single- and multi-objective evolutionary design optimization assisted
  by {G}aussian random field metamodels, 2005.
\newblock PhD thesis, FB Informatik, University of Dortmund.

\bibitem[Eric et~al.(2007)Eric, Freitas, and Ghosh]{eric2007active}
B.~Eric, N.~Freitas, and A.~Ghosh.
\newblock Active preference learning with discrete choice data.
\newblock \emph{Advances in neural information processing systems}, 20, 2007.

\bibitem[Fonseca and Fleming(1995)]{fonseca1995overview}
C.~M. Fonseca and P.~J. Fleming.
\newblock An overview of evolutionary algorithms in multiobjective
  optimization.
\newblock \emph{Evolutionary computation}, 3\penalty0 (1):\penalty0 1--16,
  1995.

\bibitem[Giagkiozis and Fleming(2015)]{chebyshev}
I.~Giagkiozis and P.~J. Fleming.
\newblock Methods for multi-objective optimization: An analysis.
\newblock \emph{Information Sciences}, 293:\penalty0 338--350, 2015.

\bibitem[Gonz{\'a}lez et~al.(2017)Gonz{\'a}lez, Dai, Damianou, and
  Lawrence]{gonzalez2017preferential}
J.~Gonz{\'a}lez, Z.~Dai, A.~Damianou, and N.~D. Lawrence.
\newblock Preferential bayesian optimization.
\newblock In \emph{International Conference on Machine Learning}, pages
  1282--1291. PMLR, 2017.

\bibitem[Hakanen and Knowles(2017)]{hakanen2017using}
J.~Hakanen and J.~D. Knowles.
\newblock On using decision maker preferences with {ParEGO}.
\newblock In H.~Trautmann, G.~Rudolph, K.~Klamroth, O.~Sch{\"u}tze, M.~Wiecek,
  Y.~Jin, and C.~Grimme, editors, \emph{Evolutionary Multi-Criterion
  Optimization}, pages 282--297, Cham, 2017. Springer International Publishing.

\bibitem[Hakanen et~al.(2016)Hakanen, Chugh, Sindhya, Jin, and
  Miettinen]{hakanen2016connections}
J.~Hakanen, T.~Chugh, K.~Sindhya, Y.~Jin, and K.~Miettinen.
\newblock Connections of reference vectors and different types of preference
  information in interactive multiobjective evolutionary algorithms.
\newblock In \emph{2016 IEEE Symposium Series on Computational Intelligence
  (SSCI)}, pages 1--8, 2016.

\bibitem[He et~al.(2016)He, Zhang, Ren, and Sun]{resnet}
K.~He, X.~Zhang, S.~Ren, and J.~Sun.
\newblock Deep residual learning for image recognition.
\newblock In \emph{Proceedings of the IEEE conference on computer vision and
  pattern recognition}, pages 770--778, 2016.

\bibitem[He et~al.(2020)He, Sun, Song, Wang, and Usmani]{he2020preference}
Y.~He, J.~Sun, P.~Song, X.~Wang, and A.~S. Usmani.
\newblock Preference-driven kriging-based multiobjective optimization method
  with a novel multipoint infill criterion and application to airfoil shape
  design.
\newblock \emph{Aerospace Science and Technology}, 96:\penalty0 105555, 2020.

\bibitem[Hern{\'a}ndez-Lobato et~al.(2016)Hern{\'a}ndez-Lobato,
  Hernandez-Lobato, Shah, and Adams]{hernandez2016predictive}
D.~Hern{\'a}ndez-Lobato, J.~Hernandez-Lobato, A.~Shah, and R.~Adams.
\newblock Predictive entropy search for multi-objective bayesian optimization.
\newblock In \emph{International conference on machine learning}, pages
  1492--1501. PMLR, 2016.

\bibitem[Hern\'{a}ndez-Lobato et~al.(2014)Hern\'{a}ndez-Lobato, Hoffman, and
  Ghahramani]{Hernandez2014-Predictive}
J.~M. Hern\'{a}ndez-Lobato, M.~W. Hoffman, and Z.~Ghahramani.
\newblock Predictive entropy search for efficient global optimization of
  black-box functions.
\newblock In \emph{Advances in Neural Information Processing Systems 27}, pages
  918--926. Curran Associates, Inc., 2014.

\bibitem[Houlsby et~al.(2011)Houlsby, Husz{\'a}r, Ghahramani, and
  Lengyel]{houlsby2011bayesian}
N.~Houlsby, F.~Husz{\'a}r, Z.~Ghahramani, and M.~Lengyel.
\newblock Bayesian active learning for classification and preference learning,
  2011.

\bibitem[Ishibuchi et~al.(2008)Ishibuchi, Tsukamoto, and
  Nojima]{ishibuchi2008evolutionary}
H.~Ishibuchi, N.~Tsukamoto, and Y.~Nojima.
\newblock Evolutionary many-objective optimization: A short review.
\newblock In \emph{2008 IEEE Congress on Evolutionary Computation (IEEE World
  Congress on Computational Intelligence)}, pages 2419--2426, 2008.

\bibitem[Jerry~Lin et~al.(2022)Jerry~Lin, Astudillo, Frazier, and
  Bakshy]{lin2022preference}
Z.~Jerry~Lin, R.~Astudillo, P.~Frazier, and E.~Bakshy.
\newblock Preference exploration for efficient bayesian optimization with
  multiple outcomes.
\newblock In G.~Camps-Valls, F.~J.~R. Ruiz, and I.~Valera, editors,
  \emph{Proceedings of The 25th International Conference on Artificial
  Intelligence and Statistics}, volume 151 of \emph{Proceedings of Machine
  Learning Research}, pages 4235--4258. PMLR, 2022.

\bibitem[Ke et~al.(2017)Ke, Meng, Finley, Wang, Chen, Ma, Ye, and
  Liu]{ke2017lightgbm}
G.~Ke, Q.~Meng, T.~Finley, T.~Wang, W.~Chen, W.~Ma, Q.~Ye, and T.-Y. Liu.
\newblock Lightgbm: A highly efficient gradient boosting decision tree.
\newblock \emph{Advances in neural information processing systems}, 30, 2017.

\bibitem[Knowles(2006)]{knowles2006parego}
J.~Knowles.
\newblock Parego: A hybrid algorithm with on-line landscape approximation for
  expensive multiobjective optimization problems.
\newblock \emph{IEEE transactions on evolutionary computation}, 10\penalty0
  (1):\penalty0 50--66, 2006.

\bibitem[Krizhevsky et~al.(2009)Krizhevsky, Hinton, et~al.]{cifar10}
A.~Krizhevsky, G.~Hinton, et~al.
\newblock Learning multiple layers of features from tiny images.
\newblock 2009.

\bibitem[Kursawe(1990)]{kursawe1990variant}
F.~Kursawe.
\newblock A variant of evolution strategies for vector optimization.
\newblock In \emph{International conference on parallel problem solving from
  nature}, pages 193--197. Springer, 1990.

\bibitem[Li et~al.(2021)Li, Zhang, Thrampoulidis, Chen, and Oymak]{autobalance}
M.~Li, X.~Zhang, C.~Thrampoulidis, J.~Chen, and S.~Oymak.
\newblock {AutoBalance}: Optimized loss functions for imbalanced data.
\newblock \emph{Advances in Neural Information Processing Systems},
  34:\penalty0 3163--3177, 2021.

\bibitem[Minka(2001)]{minka2001expectation}
T.~P. Minka.
\newblock Expectation propagation for approximate bayesian inference.
\newblock In \emph{Proceedings of the Seventeenth Conference on Uncertainty in
  Artificial Intelligence}, page 362\UTF{2013}369, San Francisco, CA, USA,
  2001. Morgan Kaufmann Publishers Inc.
\newblock ISBN 1558608001.

\bibitem[Moro et~al.(2012)Moro, Rita, and Cortez]{misc_bank_marketing_222}
S.~Moro, P.~Rita, and P.~Cortez.
\newblock {Bank Marketing}.
\newblock UCI Machine Learning Repository, 2012.
\newblock {DOI}: https://doi.org/10.24432/C5K306.

\bibitem[Palar et~al.(2018)Palar, Yang, Shimoyama, Emmerich, and
  B\"{a}ck]{palar2018multi}
P.~S. Palar, K.~Yang, K.~Shimoyama, M.~Emmerich, and T.~B\"{a}ck.
\newblock Multi-objective aerodynamic design with user preference using
  truncated expected hypervolume improvement.
\newblock In \emph{Proceedings of the Genetic and Evolutionary Computation
  Conference}, GECCO '18, page 1333\UTF{2013}1340, New York, NY, USA, 2018.
  Association for Computing Machinery.

\bibitem[Paria et~al.(2020)Paria, Kandasamy, and P{\'o}czos]{mobors}
B.~Paria, K.~Kandasamy, and B.~P{\'o}czos.
\newblock A flexible framework for multi-objective bayesian optimization using
  random scalarizations.
\newblock In \emph{Uncertainty in Artificial Intelligence}, pages 766--776.
  PMLR, 2020.

\bibitem[Poloni et~al.(1995)]{poloni1995hybrid}
C.~Poloni et~al.
\newblock Hybrid ga for multi objective aerodynamic shape optimisation.
\newblock In \emph{Genetic algorithms in engineering and computer science},
  pages 397--415. John Wiley \& Sons Ltd, 1995.

\bibitem[Riihim{\"a}ki and Vehtari(2010)]{monotonicGPR}
J.~Riihim{\"a}ki and A.~Vehtari.
\newblock Gaussian processes with monotonicity information.
\newblock In \emph{Proceedings of the thirteenth international conference on
  artificial intelligence and statistics}, pages 645--652. JMLR Workshop and
  Conference Proceedings, 2010.

\bibitem[Schaffer et~al.(1985)]{schaffer1985multiple}
J.~D. Schaffer et~al.
\newblock Multiple objective optimization with vector evaluated genetic
  algorithms.
\newblock In \emph{Proceedings of an international conference on genetic
  algorithms and their applications}, pages 93--100, 1985.

\bibitem[Sui et~al.(2018)Sui, Zoghi, Hofmann, and Yue]{sui2018advancements}
Y.~Sui, M.~Zoghi, K.~Hofmann, and Y.~Yue.
\newblock Advancements in dueling bandits.
\newblock In \emph{Proceedings of the 27th International Joint Conference on
  Artificial Intelligence}, page 5502\UTF{2013}5510. AAAI Press, 2018.

\bibitem[Suzuki et~al.(2020)Suzuki, Takeno, Tamura, Shitara, and
  Karasuyama]{suzuki2020multi}
S.~Suzuki, S.~Takeno, T.~Tamura, K.~Shitara, and M.~Karasuyama.
\newblock Multi-objective bayesian optimization using pareto-frontier entropy.
\newblock In \emph{International Conference on Machine Learning}, pages
  9279--9288. PMLR, 2020.

\bibitem[Takeno et~al.(2023)Takeno, Nomura, and Karasuyama]{takeno2023towards}
S.~Takeno, M.~Nomura, and M.~Karasuyama.
\newblock Towards practical preferential {B}ayesian optimization with skew
  {G}aussian processes.
\newblock In \emph{Proceedings of the 40th International Conference on Machine
  Learning}, volume 202 of \emph{Proceedings of Machine Learning Research},
  pages 33516--33533. PMLR, 2023.

\bibitem[Taylor et~al.(2021)Taylor, Ha, Li, Chan, and Li]{taylor2021bayesian}
K.~Taylor, H.~Ha, M.~Li, J.~Chan, and X.~Li.
\newblock {B}ayesian preference learning for interactive multi-objective
  optimisation.
\newblock In \emph{Proceedings of the Genetic and Evolutionary Computation
  Conference}, GECCO '21, page 466\UTF{2013}475, New York, NY, USA, 2021.
  Association for Computing Machinery.

\bibitem[Williams and Rasmussen(2006)]{gp}
C.~K. Williams and C.~E. Rasmussen.
\newblock \emph{Gaussian processes for machine learning}, volume~2.
\newblock MIT press Cambridge, MA, 2006.

\bibitem[Zafar et~al.(2017)Zafar, Valera, Rogriguez, and
  Gummadi]{zafar2017fairness}
M.~B. Zafar, I.~Valera, M.~G. Rogriguez, and K.~P. Gummadi.
\newblock Fairness constraints: Mechanisms for fair classification.
\newblock In \emph{Artificial intelligence and statistics}, pages 962--970.
  PMLR, 2017.

\end{thebibliography}
\bibliographystyle{abbrvnat}

\newpage
\appendix
\onecolumn

\section{Mutual Information for IR}
\label{app:acq-ir}

We need to select 
$\*f^{i}$
that efficiently reduces the uncertainty of $\*w$.
Let 
\begin{align*}
 z_{\rm IR} = 
 \begin{cases}
  1 & \text{ if } 1 \succ \ell_i^{\prime} \text { for } \ell_i^{\prime} \in[L] \backslash 1, 
  \\
\vdots
\\
  L & \text{ if } L \succ \ell_i^{\prime} \text { for } \ell_i^{\prime} \in[L] \backslash L, 
 \end{cases}
\end{align*}
be the indicator of the preference given by the DM.
Our AL acquisition function is defined by the MI between $z_{\rm IR}$ and $\*w$:
\begin{align}
 \mathrm{MI}(z_{\rm IR} ; \*w) = H[ z_{\rm IR} ] - \EE_{\*w}[H[ z_{\rm IR} \mid \*w]].
 \label{eq:MI-IR}
\end{align}
For the first term of \eq{eq:MI-IR}, since $z_{\rm IR}$ follows the categorical distribution, we have
\begin{align*}
 H[z_{\rm IR}] 
 = \sum_{z_{\rm IR} \in \{ 1, \ldots, L \}} p(z_{\rm IR}) \log p(z_{\rm IR}).
\end{align*}
Similar to the case of PC, 
$p(z_{\rm IR})$, 
in which $\*w$ is marginalized, is difficult to evaluate analytically.
On the other hand, the conditional distribution
$p(z_{\rm IR} \mid \*w)$
is easy to evaluate as follows: 
%
\begin{align*}
 p(z_{\rm IR} \mid \*w)
 = \prod_{z_{\rm IR}^{\prime} \in \{ 1, \ldots, L \} \backslash z_{\rm IR}} p(z_{\rm IR} \succ z_{\rm IR}^{\prime} \mid \*w).
\end{align*}
Therefore, we employ a sampling based approximation
\begin{align*}
 p(z_{\rm IR}) \approx \sum_{\*w \in \cW} p(z_{\rm IR} \mid \*w) / |\cW|,
\end{align*}
where $\cW$ is a set of $\*w$ generated from the posterior.
%
For the second term of \eq{eq:MI-IR}, the same sample set $\cW$ can be used as
\begin{align*}
 & \EE_{\*w}[H[ z_{\rm IR} \mid \*w]] \approx 
 \\
 & \ \sum_{\*w \in \cW} \sum_{z_{\rm IR} \in \{ 1, \ldots, L \}} p(z_{\rm IR} \mid \*w) \log p(z_{\rm IR} \mid \*w) / |\cW|.
\end{align*}
%


\section{Expectation Propagation for Preferential GP with Monotonicity Constraint}
\label{app:pregpmono}

We let the approximated posterior as follows:
\begin{align*}
q(\*U, \*U^{\prime} | D_{\rm{pre}}, \*m) \propto p(\*U, \*U^{\prime}) \prod^n_{j=1} \tilde{t}_j(\*U_j;\tilde{s}_j, \tilde{\*m}_j, \tilde{\*\pi}_j)\prod^M_{i=1} \tilde{u}_i(\tilde{Z}_i, \tilde{\mu}_i, \tilde{\sigma}_i^2),
\end{align*}
where $\*U_j=\left(U(\*f^{(j)}), U(\*f^{(j^{\prime})})\right)^{\top}$, $\tilde{t}_j(\*U_j;\tilde{s}_j, \tilde{\*m}_j, \tilde{\*\pi}_j)=\tilde{s}_j\exp\left\{\frac{1}{2}(\*U_j - \tilde{\*m}_j)^{\top}\tilde{\*\pi}_j^{-1}(\*U_j - \tilde{\*m}_j)\right\}$, and $\tilde{u}_i(\tilde{Z}_i, \tilde{\mu}_i, \tilde{\sigma}_i^2) = \tilde{Z}_i\mathcal{N}(\tilde{\mu}_i, \tilde{\sigma}_i^2)$. Then, $q(\*U, \*U^{\prime} | D_{\rm{pre}}, \*m)$ follows multi-variate normal distribution which has the mean $\*\mu = \*\Sigma\tilde{\*\Sigma}_{\rm joint}^{-1}\tilde{\*\mu}_{\rm joint}$ and the covariance matrix $\*\Sigma = (\*K_{\rm joint}^{-1} + \tilde{\*\Sigma}_{\rm joint}^{-1})^{-1}$, where
\begin{align}
\tilde{\*\mu}_{\rm joint} = 
\begin{bmatrix}
\*M\\
\tilde{\*\mu}
\end{bmatrix}
, \tilde{\*\Sigma}_{\rm joint}=
\begin{bmatrix}
\*\Pi^{-1} & \*0\\
\*0 & \tilde{\*\Sigma}
\end{bmatrix}.
\label{eq:ep-mean-cov}
\end{align}
In \eq{eq:ep-mean-cov}, $\tilde{\*\mu}$ is the vector of site means $\tilde{\mu}_i$, $\tilde{\*\Sigma}$ is a diagonal matrix with site variances $\tilde{\sigma}_i^2$ on the diagonal, $\*\Pi =\sum_{j=1}^n \*\Pi_j$ ($\*\Pi_j$ is an augmented matrix for $\tilde{\*\pi}_j$, which is a $n \times n$ matrix with only four non-zero entries from $\tilde{\*\pi}_j$), and $\*M = \sum_{j=1}^{n} \tilde{\*M}_j$ ($\tilde{\*M}_j$ is a $n \times 1$ vector for $\tilde{\*m}_j$).
Here, $\*U_j, \tilde{s}_j, \tilde{\*m}_j, \tilde{\*\pi}_j$ and $\tilde{Z}_i, \tilde{\mu}_i, \tilde{\sigma}_i^2$ are the parameters to update by EP.
The detailed update equations of them are in \citep{chu2005extensions} and \citep{monotonicGPR}.
%

%

\section{Numerical Calculation of EI for CSF}
\label{app:ei-calculation}

In our EI \eqref{eq:EI}, the joint expectation over $\*f(\*x)$ and $\*w$ are required to evaluate (i.e., $\EE_{\*f(\*x), \*w}$).
For the case of CSF, the expectation over $\*f(\*x)$ can be transformed into an integration over the one dimensional space of $U(\*f)$, by which general numerical integration methods are easily applicable instead of the MC method on the $L$ dimensional space of $\*f(\*x)$.
%
The EI acquisition function \eqref{eq:EI} can be rewritten as follows:
\begin{align}
 \alpha_{\rm EI}(\*x) &= 
 \EE_{\*f(\*x), \*w}\left[ \max \left\{
 U \left( \*f(\*x) \right) 
 - 
 U_{\rm best}, 
 0
 \right\} \right]
 \notag \\
 &= \int_{\*w}p(\*w\mid D_{\rm pre}) \int_{\*f(\*x)} p( \*f(\*x) \mid \*w )
 \max \left\{ U \left( \*f(\*x) \right) - U_{\rm best}, 0 \right\}
 \mathrm{d} \*f(\*x) \mathrm{d} \*w 
 \notag \\
 &= \int_{\*w}p(\*w\mid D_{\rm pre}) 
 \int^{\infty}_{U_{\rm best}} 
 h(u \mid \*w)\left(u - 
 U_{\rm best}
 \right)\mathrm{d} u \mathrm{d} \*w, 
 \label{eq:EI-WU}
\end{align}
where $u = U\left(\*f(\*x)\right) = \min\left(\frac{f_1(\*x)}{w_1},\ldots, \frac{f_L(\*x)}{w_L}\right)$ and $h(u \mid \*w)$ is the probability density function of $u$ given $\*w$.
The last line of the above equations \eqref{eq:EI-WU} is from the law of unconscious statistician. 
The cumulative distribution function (CDF) of $h(u \mid \*w)$ is
\begin{align*}
H(u \mid \*w) &= 1 - \prod_{i=1}^L p\left(\frac{f_i(\*x)}{w_i} \geq u \ \Bigg| \ \*w \right)
\\
&= 1 - \prod_{i=1}^L \left(1 - p\left(\frac{f_i(\*x)}{w_i} < u \ \Bigg| \ \*w \right)\right)
\\
&= 1 - \prod_{i=1}^L \left(1 - \Phi\left(\frac{w_iu - \mu_i(\*x)}{\sigma_i(\*x)}\right)\right),
\end{align*}
where $\mu_i(\*x)$ and $\sigma_i(\*x)$ are the mean and the standard deviation of the GP posterior of $f_i(\*x)$, respectively.
We obtain $h(u \mid \*w)$ as the derivative of $H(u \mid \*w)$.
\begin{align*}
h(u \mid \*w) &= \frac{d}{du}H(u \mid \*w)\\
&= \sum_{j=1}^L\left(\left(\frac{d}{du}\left(1-\Phi\left(\frac{w_ju - \mu_j(\*x)}{\sigma_j(\*x)}\right)\right)\right)\prod_{i=1, i\neq j}^L \left(1 - \Phi\left(\frac{w_iu - \mu_i(\*x)}{\sigma_i(\*x)}\right)\right)\right)\\
&= \sum_{j=1}^L\left(-\frac{w_j}{\sigma_j(\*x)}\phi\left(\frac{w_ju - \mu_j(\*x)}{\sigma_j(\*x)}\right)\prod_{i=1, i\neq j}^L \left(1 - \Phi\left(\frac{w_iu - \mu_i(\*x)}{\sigma_i(\*x)}\right)\right)\right),
\end{align*}
where $\phi$ is the probability density function of the standard normal distribution.
For the inner integral in \eqref{eq:EI-WU}, i.e., 
$\int^{\infty}_{U_{\rm best}} h(u \mid \*w)\left(u - U_{\rm best} \right) \mathrm{d} u$, 
it is easy to apply general numerical integration methods (such as functions in the python {\tt scipy.integrate} library) because the integrand is easy to calculate using the above $h(u \mid \*w)$.
%
%

\section{Settings of BO}
\label{app:setting-bo}

BO is implemented by the python package GPy. 
For GPs of $f_{\ell}(\*\theta)$, the RBF kernel 
$k\left(\*x, \*x^{\prime}\right)=\theta_1 \exp \left(-\frac{\left\|\*x-\*x^{\prime}\right\|^2}{\theta_2}\right)$
is used, where $\theta_1$ and $\theta_2$ are hyper-parameters.
The hyper-parameters of the kernel were optimized by the marginal likelihood maximization. 
The objective function value was scaled in $[0,1]$.
The number of initial points was 4 (randomly selected).

\section{Additional Information and Results on Experiments for Benchmark Function}

\subsection{Details of Benchmark Function}
\label{app:setting-bench}

\subsubsection{DTLZ1}

The DTLZ1 function, in which we set the input dimension $d = 3$ and the output dimension $L = 3$, is defined as follows:
\begin{align*}
\begin{split}
f_1(\*{x})&=\frac{1}{2} x_1 x_2 \cdots x_{L-1}\left(1+g\left(\*{x}_L\right)\right) \\
f_2(\*{x})&=\frac{1}{2} x_1 x_2 \cdots\left(1-x_{L-1}\right)\left(1+g\left(\*{x}_L\right)\right) \\
&\vdots \\
f_{L-1}(\*{x})&=\frac{1}{2} x_1\left(1-x_2\right)\left(1+g\left(\*{x}_L\right)\right) \\
f_L(\*{x})&=\frac{1}{2}\left(1-x_1\right)\left(1+g\left(\*{x}_L\right)\right) \\
\text { subject to } & \quad 0 \leq x_i \leq 1,  \quad\text { for } i=1,2, \ldots, d
\end{split}
\end{align*}
where
\begin{align*}
g\left(\*{x}_L\right)=100\left[\left\|\*{x}_L\right\|+\sum_{x_i \in \*{x}_L}\left(x_i-0.5\right)^2-\cos \left(20 \pi\left(x_i-0.5\right)\right)\right], 
\end{align*}
and $\*x_L$ is a subvector consisting of the last $(d - L + 1)$ elements of $\*x$. 

\subsubsection{DTLZ3}

The DTLZ3 function, in which we set the input dimension $d = 3$ and the output dimension $L = 3$, is defined as follows:
\begin{align*}
\begin{split}
&f_1(\*{x})=\left(1+g\left(\*{x}_L\right)\right)\cos\left(x_1\frac{\pi}{2}\right) \cdots  \cos\left(x_{L-2}\frac{\pi}{2}\right)\cos\left(x_{L-1}\frac{\pi}{2}\right) \\
&f_2(\*{x})=\left(1+g\left(\*{x}_L\right)\right)\cos\left(x_1\frac{\pi}{2}\right) \cdots  \cos\left(x_{L-2}\frac{\pi}{2}\right)\sin\left(x_{L-1}\frac{\pi}{2}\right)\\
&\vdots \\
&f_{L-1}(\*{x})=\left(1+g\left(\*{x}_L\right)\right)\cos\left(x_1\frac{\pi}{2}\right) \sin\left(x_2\frac{\pi}{2}\right) \\
&f_L(\*{x})=\left(1+g\left(\*{x}_L\right)\right)\sin\left(x_1\frac{\pi}{2}\right) \\
&\text { subject to }  \quad 0 \leq x_i \leq 1,  \quad\text { for } i=1,2, \ldots, d
\end{split}
\end{align*}
where
\begin{align*}
g\left(\*{x}_L\right)=100\left[\left\|\*{x}_L\right\|+\sum_{x_i \in \*{x}_L}\left(x_i-0.5\right)^2-\cos \left(20 \pi\left(x_i-0.5\right)\right)\right], 
\end{align*}
and $\*x_L$ is a subvector consisting of the last $(d - L + 1)$ elements of $\*x$. 

\subsubsection{Kursawe}

The Kursawe function, in which we set the input dimension $d = 3$, is defined as follows:
\begin{align*}
f_1(\*{x})&=\sum_{i=1}^{n-1} -10\exp \left(-0.2\sqrt{x_i^2+x_{i+1}^2}\right) \\
f_2(\*{x})&=\sum_{i=1}^n |x_i|^{0.8}+5\sin \left(x_i^3\right)\\
\text { subject to }  &\quad -5 \leq x_i \leq 5,  \quad\text { for } i=1,2, \ldots, d.
\end{align*}

\subsubsection{Schaffer2}

The Schaffer2 function is defined as follows:
\begin{align*}
f_1(\*{x})&=\left\{
\begin{array}{ll}
-x & (x \leq 1)\\
x-2 & (1<x\leq3)\\
4-x & (3<x\leq4)\\
x-4 & (x>4)
\end{array}
\right.\\
f_2(\*{x})&=(x-5)^2\\
\text { subject to }  &\quad -5 \leq x \leq 10.
\end{align*}

\subsubsection{Fonseca-Fleming}
The Fonseca-Fleming function, in which we set the input dimension $d = 2$, is defined as follows:
\begin{align*}
f_1(\*{x})&=1-\exp \left(-\sum_{i=1}^n \left(x_i-\frac{1}{\sqrt{n}}\right)^2\right)\\
f_2(\*{x})&=1-\exp \left(-\sum_{i=1}^n \left(x_i+\frac{1}{\sqrt{n}}\right)^2\right)\\
\text { subject to }  &\quad -4 \leq x_i \leq 4,  \quad\text { for } i=1,2, \ldots, d.
\end{align*}

\subsubsection{Poloni}
The Poloni function is defined as follows:
\begin{align*}
&f_1(\*x)=\left[1+\left(A_1-B_1(\*x)\right)^2+\left(A_2-B_2(\*x)\right)^2\right] \\
&f_2(\*x)=(x_1+3)^2+(x_2+1)^2\\
&\text { where }\\
&A_1=0.5 \sin (1)-2 \cos (1)+\sin (2)-1.5 \cos (2) \\
&A_2=1.5 \sin (1)-\cos (1)+2 \sin (2)-0.5 \cos (2) \\
&B_1(\*x)=0.5 \sin (x_1)-2 \cos (x_1)+\sin (x_2)-1.5 \cos (x_2) \\
&B_2(\*x)=1.5 \sin (x_1)-\cos (x_1)+2 \sin (x_2)-0.5 \cos (x_2)\\
&\text { subject to }  \quad -\pi \leq x_1, x_2 \leq \pi.
\end{align*}

\subsubsection{Schaffer1}
The Schaffer1 function is defined as follows:
\begin{align*}
f_1(\*{x})&=x^2\\
f_2(\*{x})&=(x-2)^2\\
\text { subject to }  &\quad -10 \leq x \leq 10.
\end{align*}

\subsection{Additional Results on Benchmark Functions}
\label{app:additional-bench}

\begin{figure}
 \begin{center}
  \igr{.4}{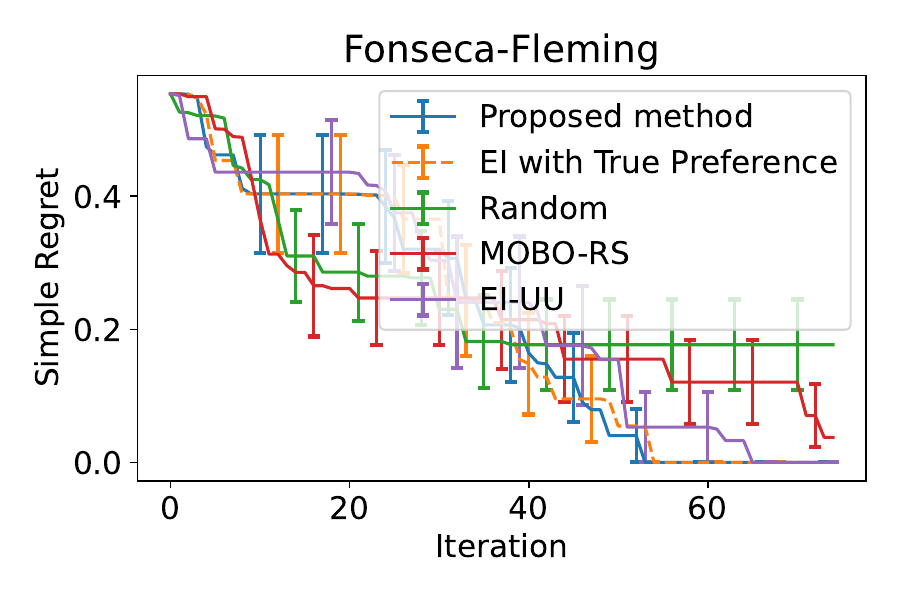}
  \igr{.4}{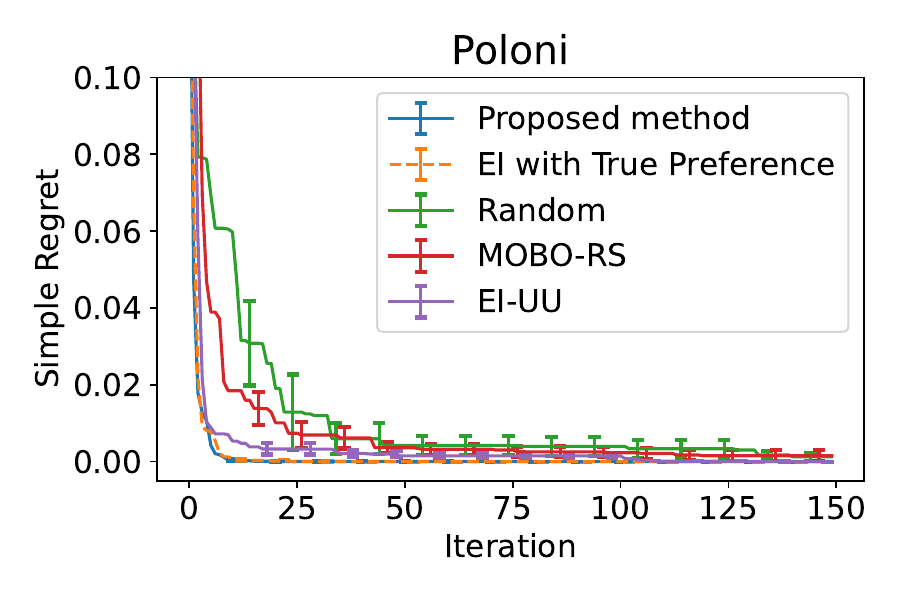}
  \igr{.4}{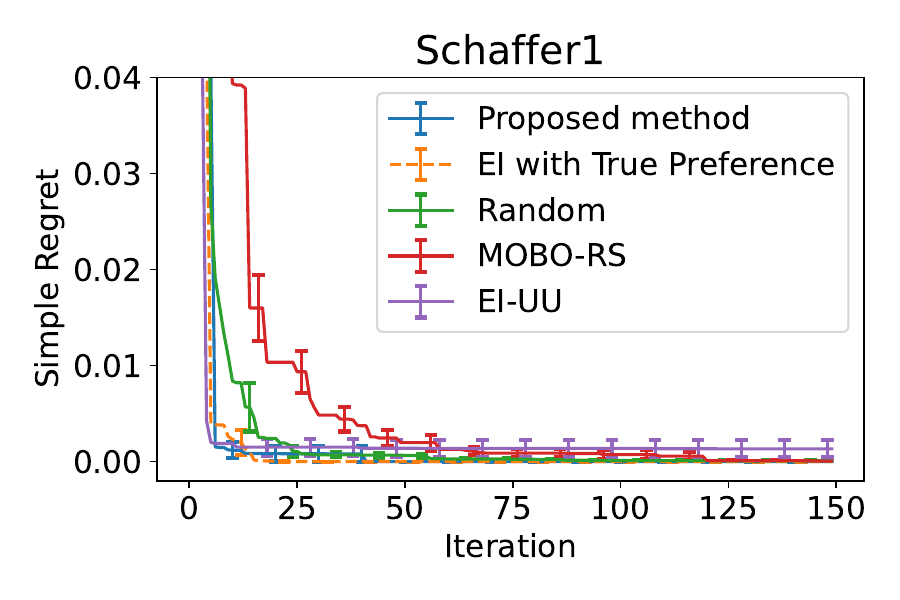}
 \end{center}
 \caption{Simple regret on additional benchmark functions.}
 \label{fig:additional-benchmark}
\end{figure}

We show additional results of BO for other well-known benchmark functions: Fonseca-Fleming \citep{fonseca1995overview}, Poloni \citep{poloni1995hybrid}, and Schaffer1 \citep{schaffer1985multiple}.
For Fonseca-Fleming, Poloni, and Schaffer1, we prepare $100$, $400$, and $1000$ input candidates, respectively (see Appendix~\ref{app:setting-bench} for detail of benchmark functions).
The results are shown in \figurename~\ref{fig:additional-benchmark}. 
We can see our proposed method outperforms Random, MOBO-RS, and EI-UU, and is comparable with EI-TP. 
%

\section{Ablation Study and Sensitivity Analysis}

\begin{figure}
 \begin{center}
  \igr{.4}{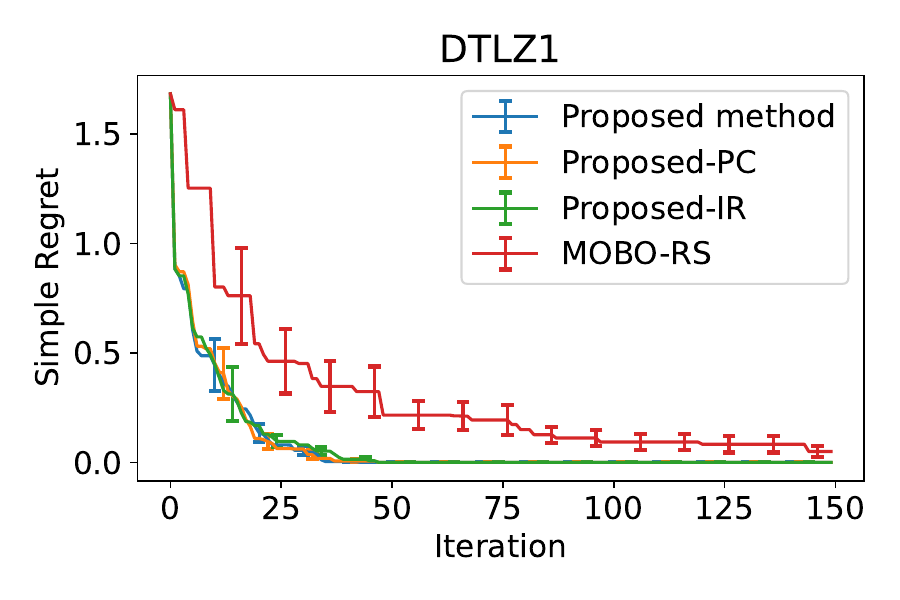}
   \igr{.4}{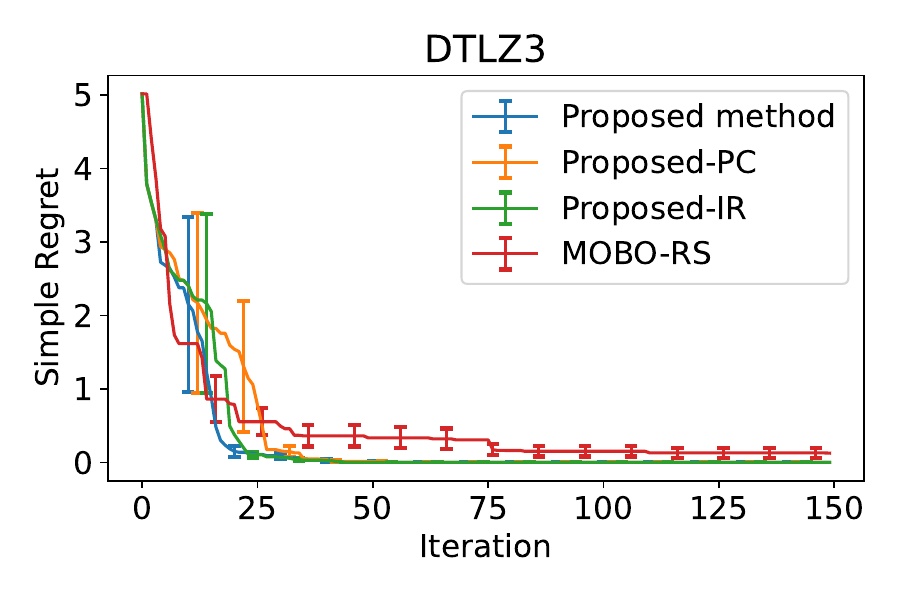}
 \end{center}
 \caption{Simple regret on benchmark functions for the ablation study on PC and IR.}
 \label{fig:ablation-study}
\end{figure}

\begin{figure}
 \begin{center}
  \igr{.4}{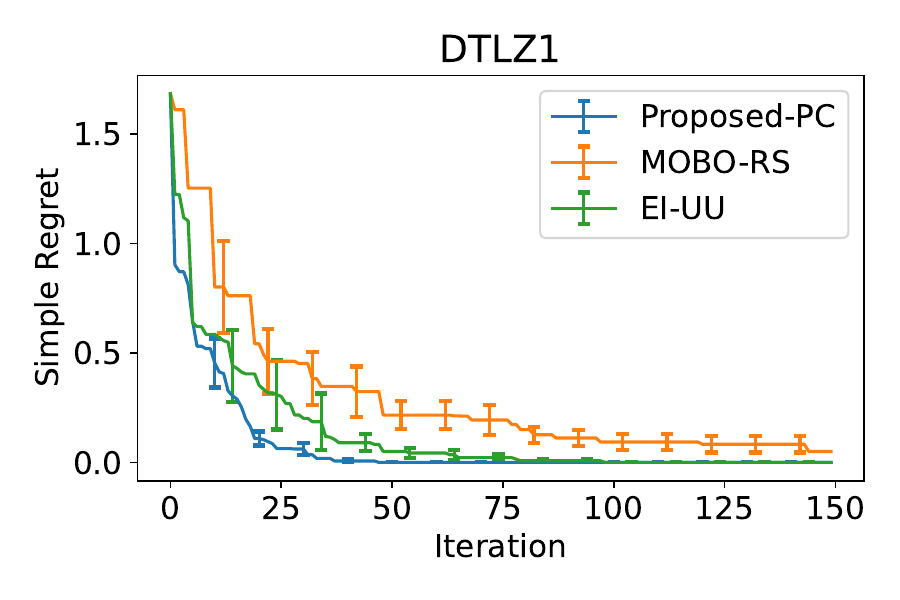}
  \igr{.4}{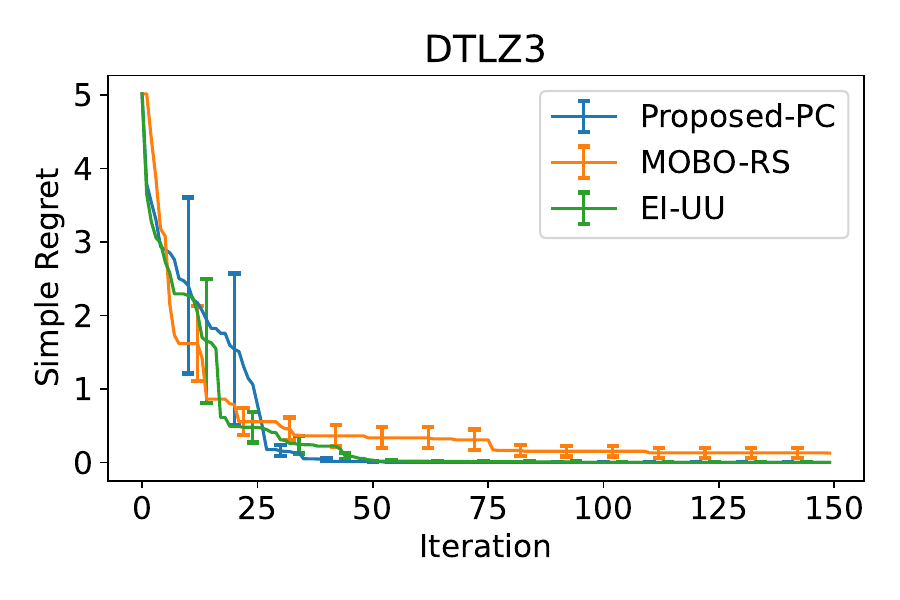}
 \end{center}
 \caption{Simple regret on benchmark functions for the comparison between the proposed method only with PC (Proposed-PC) and EI-UU.}
 \label{fig:ei-pc-ei-uu}
\end{figure}

\begin{figure}
 \begin{center}
  \igr{.4}{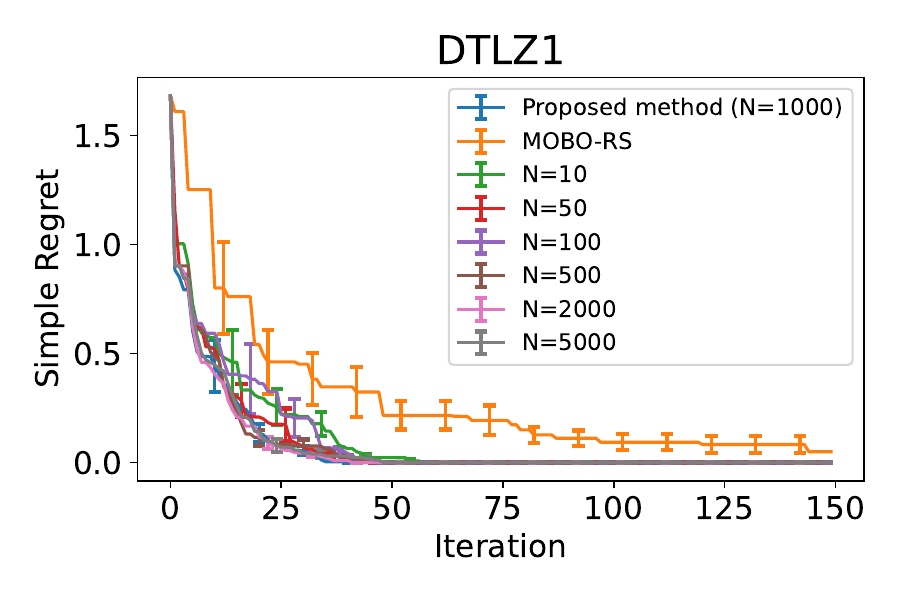}
  \igr{.4}{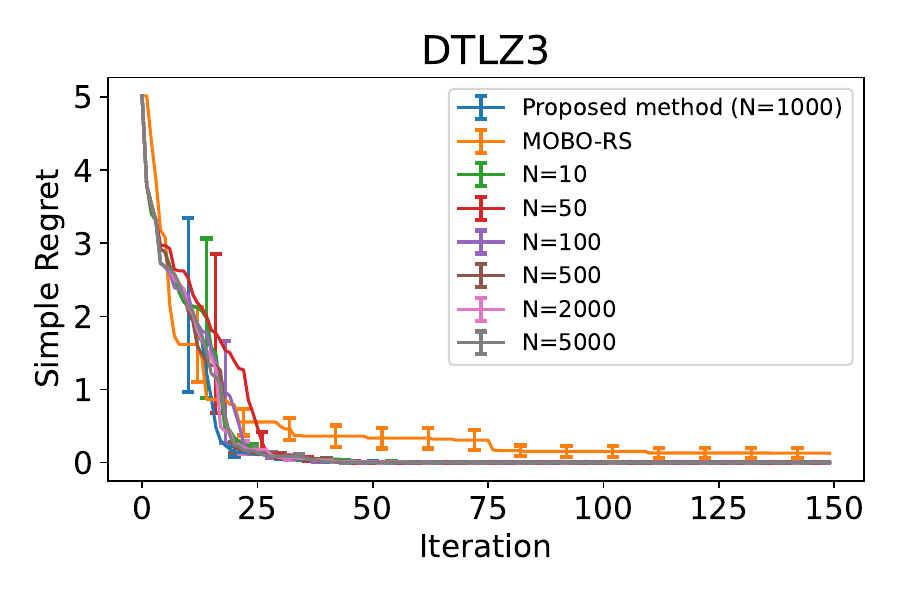}
 \end{center}
 \caption{Simple regret on benchmark functions for the sensitivity analysis.}
 \label{fig:sensitivity-analysis}
\end{figure}

\subsection{Ablation Study on PC and IR}
\label{app:ablation}

We consider two types of preference observations, i.e., PC and IR.
In this section, we evaluate how the two types of observations are effective through experiments in which only one of PC and IR is used.
We use the same benchmark functions and BO settings as before.
%
The proposed method obtains a PC- and an IR- observation at every iteration, but here we consider the proposed method only with a PC observation at every iteration, denoted as Proposed-PC, and only with an IR observation at every iteration, denoted as Proposed-IR.
%

The results for DTLZ1 and DTLZ3 are shown in \figurename~\ref{fig:ablation-study}.
%
We see that Proposed-PC and Proposed-IR can efficiently reduced simple regret compared with MOBO-RS.
%
Furthermore, the proposed method using both PC and IR is more efficient than Proposed-PC and Proposed-IR in DTLZ3.
This indicates both PC and IR work in our proposed framework.

\subsection{Evaluation only with PC Obervations}
\label{app:comparison-pc-uu}

In Section~\ref{ss:experiment-benchmark}, 
we show the comparison between our proposed method and EI-UU.
However, EI-UU only incorporates PC observation (does not use IR), so we show the comparison between the proposed method only with PC (Proposed-PC) and EI-UU.
The results are shown in \figurename~\ref{fig:ei-pc-ei-uu}.
We see that Proposed-PC outperforms EI-UU. 
In DTLZ3, EI-UU has smaller regret values at the beginning of the iterations, but from around the iteration 25, Proposed-PC has smaller values.
%

\subsection{Sensitivity Analysis on Sampling Approximation of EI}
\label{app:sensitivity}

Our proposed method calculates the values of expected improvement acquisition function using samples from the posterior of the GP (for $\*f(\*x)$) and the preference model (for $\*w$ in $U$).
We use $1000$ samples in all experiments in the main paper.
In this section, we change the number of the samples $N$. 
%
The experimental settings excluding $N$ are the same as before, and we use $N=10, 50, 100, 500, 1000$ (original), $2000,$ and $5000$.

The results are shown in \figurename~\ref{fig:sensitivity-analysis}.
We see that there is no drastic difference due to $N$ though the reduction of simple regret tends to be slow when $N$ is small.

\section{Evaluation with Other Utility Functions}
\label{app:exp-other-utility}

\subsection{Experiments using Augmented CSF Utility with Reference Point}
\label{app:augmented-csf}

\begin{figure}
 \begin{center}
 \igr{.4}{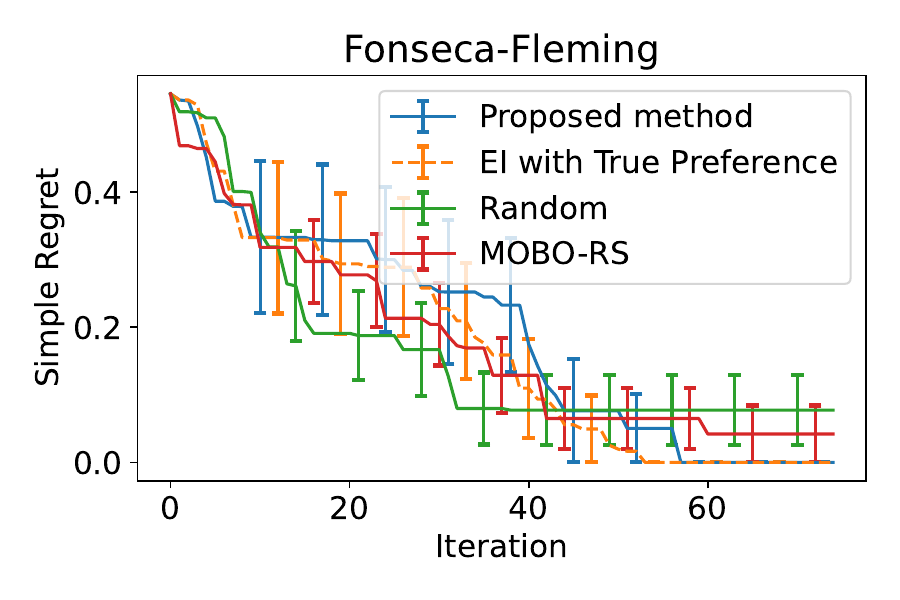}
  \igr{.4}{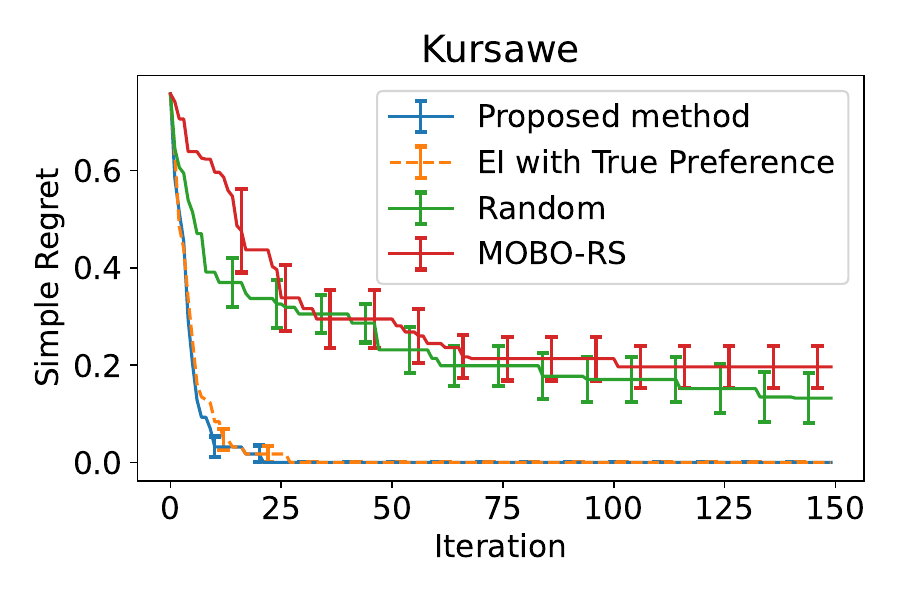}
  \igr{.4}{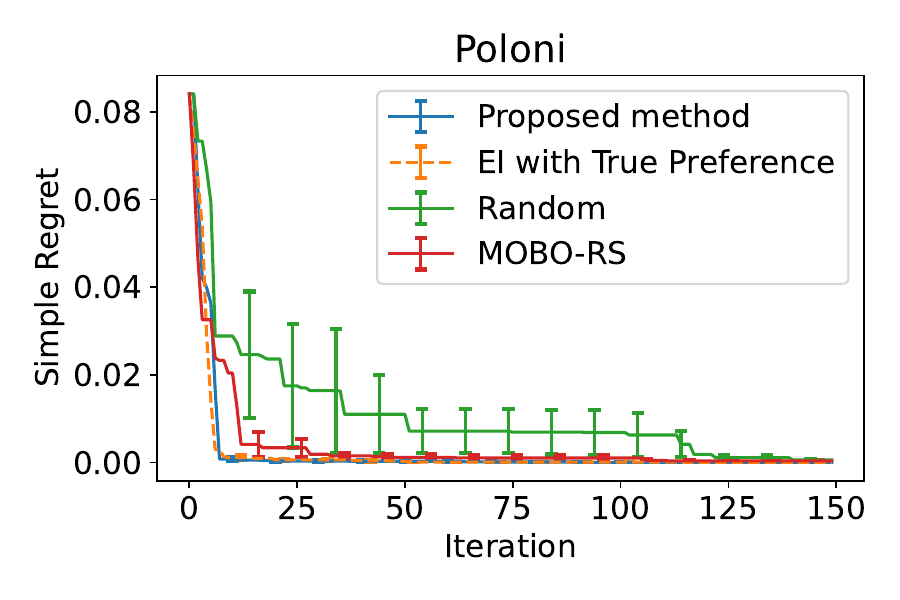}
  \igr{.4}{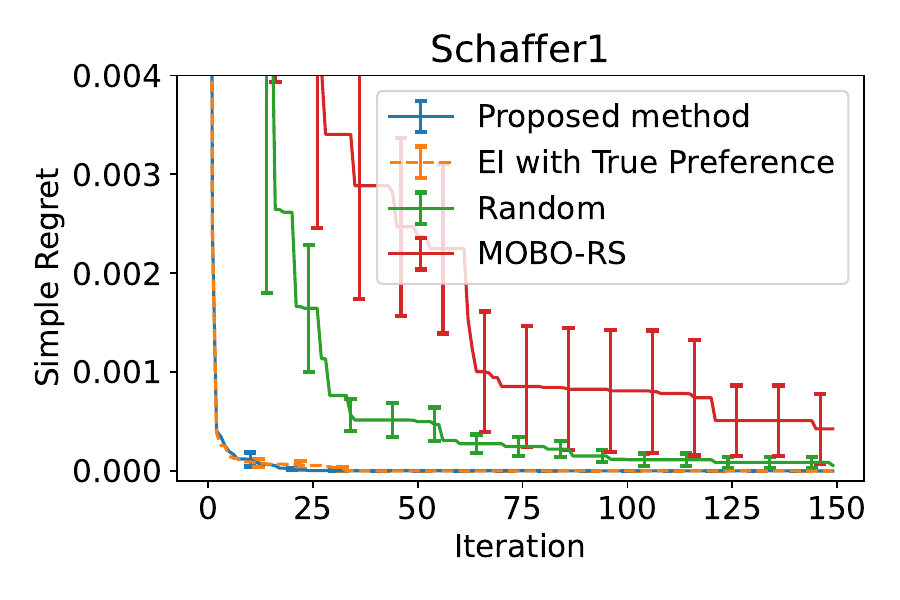}
   \igr{.4}{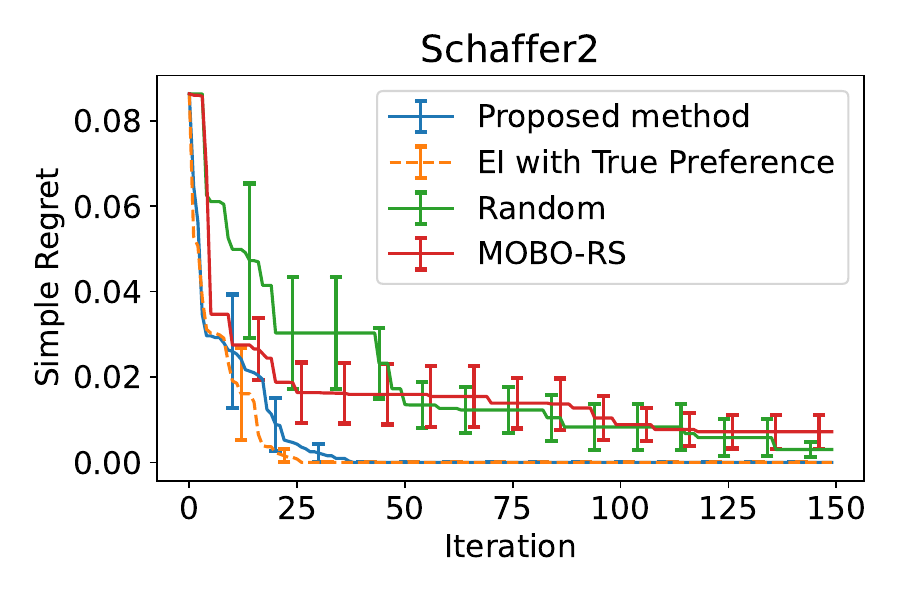}
 \end{center}
 \caption{Simple regret on benchmark functions using augmented CSF as a utility function.}
 \label{fig:augmented-csf}
\end{figure}

As we described in the paper (\S~3.5 Selection on Utility Function), 
the augmented CSF \citep{hakanen2017using}
\begin{align*}
 U(\*f(\*x)) = 
 \min_{\ell \in [L]} \frac{f_\ell(\*x) - f_\ell^{\rm ref}}{w_{\ell}} 
 + \rho \sum_{\ell \in [L]}
 \frac{f_\ell(\*x) - f_\ell^{\rm ref}}{w_{\ell}}
\end{align*}
is one of the options as our utility function.
In this section, we show additional experiments using the augmented CSF as the utility function of the proposed method. 
Here, not only $\*w$ but also $\*f^{\rm ref}=(f_1^{\rm ref},\ldots,f_L^{\rm ref})^{\top}$ is the parameter to be estimated.
Note that \citet{hakanen2017using} use this function in a different way. 
They fix $\*w$ heuristically, and $\*f^{\rm ref}$ is sampled from a box region given by the DM as the preference information \citep[see][for detail]{hakanen2017using}. 
However, specifying the box region may not be easy for the DM (the DM should provide $L \times 2$ values to define the box though these values are not necessarily easy to interpret). 
Instead, we here estimate $\*w$ and $\*f^{\rm ref}$ from PC- and IR- observations\footnote{
Unfortunately, for the estimation of the posterior of $\*f^{\rm ref}$, only PC is informative and IR cannot contribute to the estimation. 
The likelihood of IR (4) in the main text does not depend on $\*f^{\rm ref}$ because 
$g_{\ell_i}(\*f^{(i)})$
and
$g_{\ell^\prime_i}(\*f^{(i)})$
do not depend on $\*f^{\rm ref}$ (removed when taking the derivative $g_\ell(\*f) = \pd{U(\*f)}{ f^\ell}$).
As a result, IR does not have any effect on the posterior of $\*f^{\rm ref}$.
}.
%
We employ the multivariate normal distribution $\mathcal{N}(\*0, \frac{1}{100}I)$ as an example of the prior distribution of $\*f^{\rm ref}$.
The true utility function (ground truth) is also expressed as the augmented CSF in this experiment, and the parameter $\*w_{\rm true}$ determined by sampling from the Dirichlet distribution ($\*\alpha = (2, \ldots, 2)^{\top}$) and $\*f^{\rm ref}$ is sampling from multivariate normal distribution $\mathcal{N}(\*0, \frac{1}{100}I)$.
The augmentation coefficient $\rho$ is set as $0.001$.

The results are shown in \figurename~\ref{fig:augmented-csf}.
%
%
The proposed method reduces the simple regret efficiently compared with other methods without the preference information, and is comparable with EI-TP particularly in Kursawe, Poloni, Schaffer1, and Schaffer2.

\subsection{Additional Results with GP-based Utility Function}

\begin{figure}
 \begin{center}
  \igr{.4}{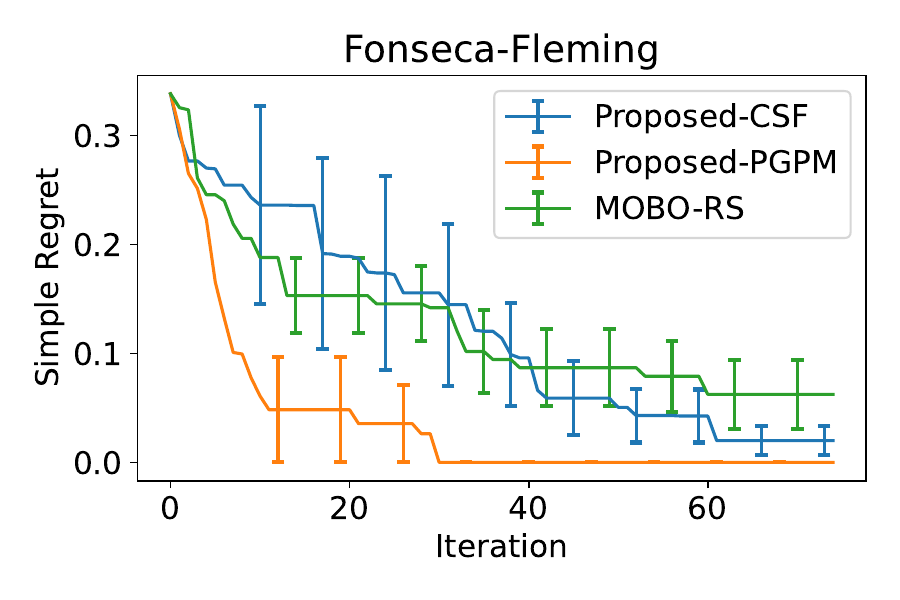}
  \igr{.4}{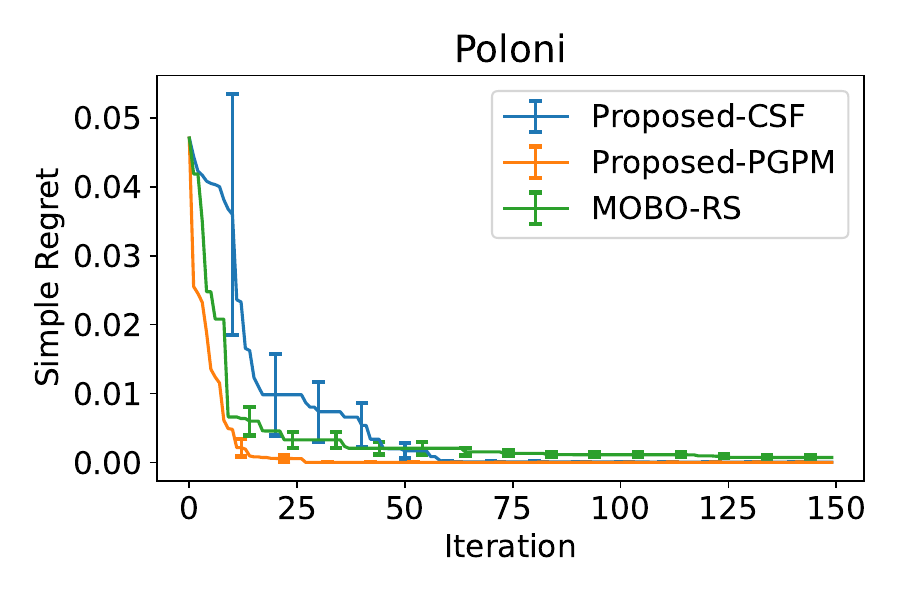}
  \igr{.4}{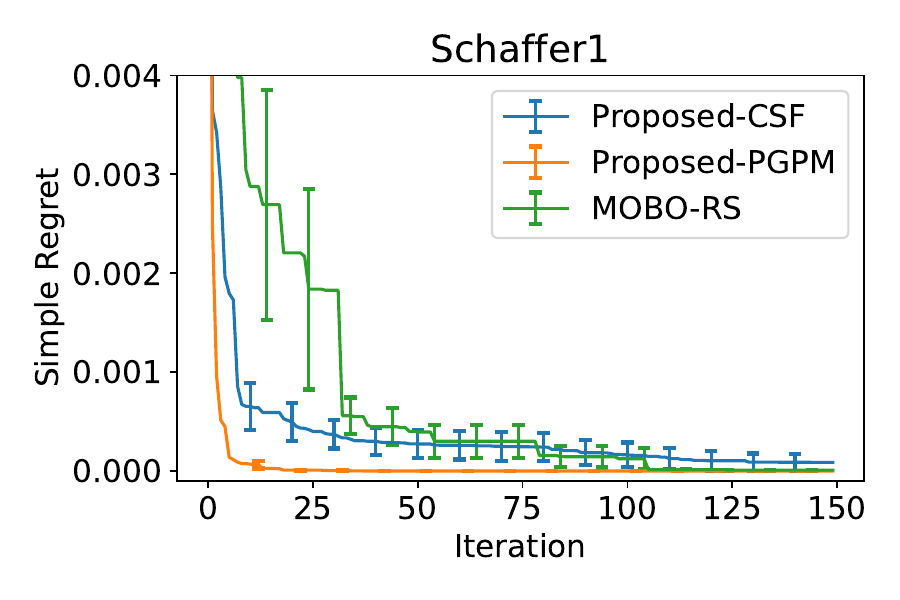}
 \end{center}
 \caption{Simple regret on benchmark functions using the CSF model and the GP-based model.}
 \label{fig:preferential-gp-app}
\end{figure}

Results on benchmark functions called 
Fonseca-Fleming \citep{fonseca1995overview}, 
Poloni \citep{poloni1995hybrid}, and Schaffer1 \citep{schaffer1985multiple} are shown in \figurename~\ref{fig:preferential-gp-app}. 
%
%
We see that in these three benchmark functions, Proposed-PGPM shows superior performance compared with Proposed-CSF and MOBO-RS.

%

\section{Additional Information and Results of Hyper-parameter Optimization Experiments}

We describe the details of datasets used for hyper-parameter optimization experiments for cost-sensitive learning.
Then, we show additional results of this setting.

\subsection{Details of Datasets}
\label{app:setting-hpo}

\subsubsection{Waveform-5000}

The number of classes is 3.
We divided the dataset into $\text{training}:\text{validation} = 1:1$.
The hyper-parameter $\*x \in \RR^3$ is class\_weight in LightGBM, which results in three dimensional search space. 
%
%
We generated $210$ candidate points that satisfy
$\| \*x \|_1 = 1$, 
in which values of each dimension $x_i$ are uniformly taken $20$ points from $(0, 1)$.

\subsubsection{CIFAR-10}

The number of classes is 10.
We divided the dataset into $\text{training}:\text{validation} = 5:1$.
The hyper-parameter $\*x \in \RR^{10}$ is the weights of weighted cross-entropy loss of a neural network, which results in ten dimensional search space.
%
%
We generated $5005$ candidate points that satisfy
$\| \*x \|_1 = 1$, 
in which values of each dimension $x_i$ are uniformly taken $7$ points from $(0, 1)$.
The implementation is based on Pytorch and the optimizer was Adam.
The learning rate was 0.01 and the number of epoch was 20. 
The weight decay parameter was set as $10^{-4}$.

\subsubsection{Breast Cancer}
This dataset is the binary classification.
We divided the dataset into $\text{training}:\text{validation} = 1:1$.
The class balance in the training set was $1:2$, which makes the classification difficult.
The hyper-parameter $\*x$ is scale\_pos\_weight in LightGBM. 
Only for this datasets, input dimension $d=1$ because in the case of the binary classification, LightGBM only has one weight parameter (the `scale\_pos\_weight', instead of the `class\_weight' parameter).
The candidate values are $101$ points equally taken from the log space of $[10^{-5}, 10^{5}]$.

\subsubsection{Covertype}
The number of classes is 7.
We divided the dataset into $\text{training}:\text{validation} = 3:2$.
The hyper-parameter $\*x \in \RR^{7}$ is class\_weight in LightGBM, which results in three dimensional search space. 
%
%
We generated $500$ candidate points that satisfy
$\| \*x \|_1 = 1$, 
in which values of each dimension $x_i$ are uniformly taken $10$ points from $(0, 1)$.


\subsection{Additional Experiments}
\label{app:additional-hpo}

\begin{figure}
 \begin{center}
  \igr{.4}{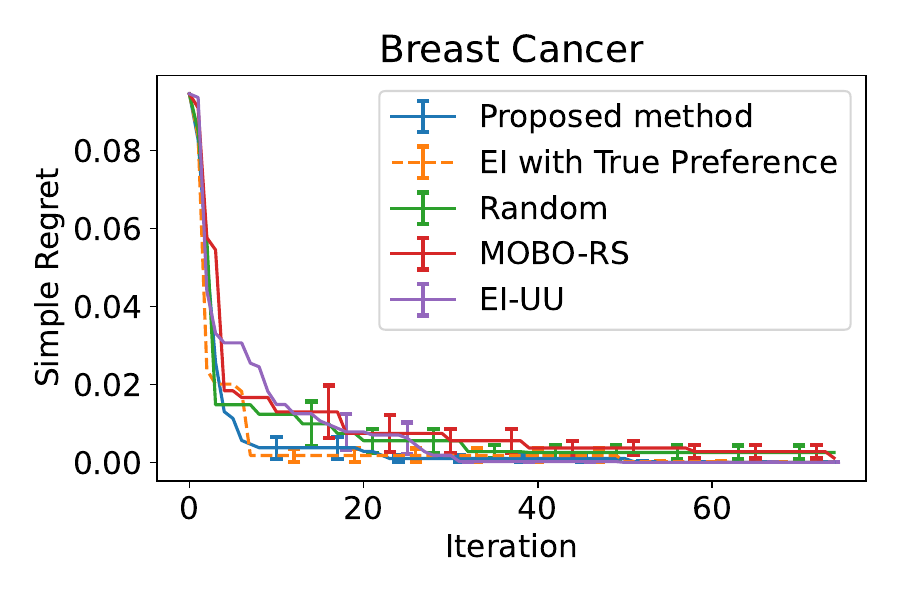}
  \igr{.4}{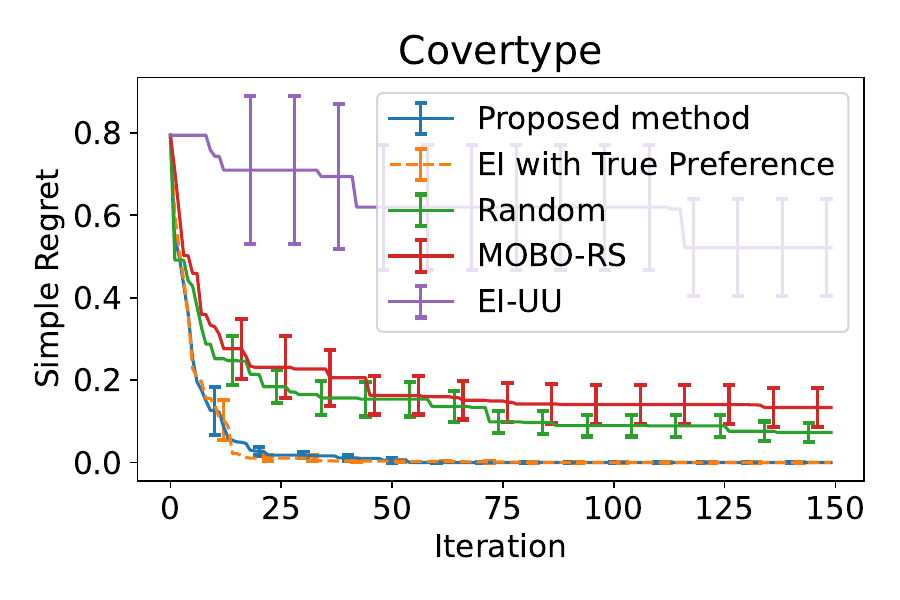}
 \end{center}
 \caption{Simple regret on additional datasets.}
 \label{fig:additional-cost-sensitive}
\end{figure}

We show additional results for other well-known datasets: Breast Cancer ($L=2$) and Covertype ($L=7$) \citep{Dua:2019}, for which details are in Appendix~\ref{app:setting-hpo}.

The results are shown in \figurename~\ref{fig:additional-cost-sensitive}. 
We can see our proposed method outperforms Random, MOBO-RS, and EI-UU, and is comparable with EI-TP. 
Details of these two datasets are as follows.

\section{Computing Infrastructure}

\begin{table}
\caption{Information of the computer with the GPU used for running experiments.}
\label{table:computer-gpu}
\centering
\begin{tabular}{c|c}\hline
CPU & Intel(R) Xeon(R) Gold 6230\\\hline
GPU & NVIDIA Quadro RTX5000\\\hline
Memory (GB) & 384\\\hline
Operating System & CentOS 7.7\\\hline
\end{tabular}
\caption{Information of the computer used for running experiments.}
\label{table:computer-cpu1}
\centering
\begin{tabular}{c|c}\hline
CPU & Intel(R) Xeon(R) Gold 6230\\\hline
Memory (GB) & 384\\\hline
Operating System & CentOS 7.7\\\hline
\end{tabular}
\caption{Information of the computer used for running experiments.}
\label{table:computer-cpu2}
\centering
\begin{tabular}{c|c}\hline
CPU & Intel(R) Xeon(R) Gold 6330\\\hline
Memory (GB) & 512\\\hline
Operating System & CentOS 7.9\\\hline
\end{tabular}
\end{table}

We show the computing infrastructure used for running experiments.
We ran neural network experiments for CIFAR-10 datasets with the computers described in \tablename~\ref{table:computer-gpu} using the GPUs.
We ran all other experiments with the computers described in \tablename~\ref{table:computer-cpu1} and \tablename~\ref{table:computer-cpu2}.
%

\end{document}